\documentclass[runningheads]{llncs}

\usepackage[T1]{fontenc}
\usepackage{graphicx}
\usepackage{enumitem}
\usepackage{cite}
\usepackage{algorithmic}
\usepackage{graphicx}
\usepackage{textcomp}
\usepackage{comment}
\usepackage{hyperref}
\usepackage{tabularx}
\usepackage{microtype}
\usepackage{pifont}
\newcommand{\cmark}{\ding{51}}%
\newcommand{\xmark}{\ding{55}}%
\usepackage{listings}
\usepackage[table]{xcolor}
\usepackage{multirow}
\usepackage{array}
\usepackage{pifont}
\usepackage{amssymb,amsfonts}

\newboolean{showcomments}
\setboolean{showcomments}{true} 
\ifthenelse{\boolean{showcomments}}
  {
		\newcommand{\nbb}[2]{
		\fcolorbox{black}{yellow}{\bfseries\sffamily\scriptsize#1}
		{\sf$\blacktriangleright$\textcolor{blue}{\textit{#2}}$\blacktriangleleft$}
		}
		
		\newcommand{\remarks}[1]{\color{red}[#1]\color{black}}

		\newcommand{\del}[1]{\textcolor{red}{\sout{#1}}} 
  }
  {
		\newcommand{\nbb}[2]{}
		\newcommand{\remarks}[1]{}

		\newcommand{\del}[1]{} 
  }

\begin{document}

\title{LLMs Can Check Their Own Results to Mitigate Hallucinations in Traffic Understanding Tasks}

\titlerunning{Mitigating Hallucinations in Traffic Understanding}

\author{Malsha Ashani Mahawatta Dona \inst{1}\orcidID{0009-0003-3812-5466} \and
Beatriz Cabrero-Daniel\inst{1}\orcidID{0000-0001-5275-8372} \and
Yinan Yu\inst{2}\orcidID{0000-0002-3221-7517} \and
Christian Berger\inst{1}\orcidID{0000-0002-4828-1150}}
\authorrunning{M.A.M Dona et al.}

\institute{
University of Gothenburg, Gothenburg, Sweden
\email{\{malsha.mahawatta,beatriz.cabrero-daniel,christian.berger\}@gu.se}\\
\and
Chalmers University of Technology, Gothenburg, Sweden\\
\email{yinan@chalmers.se}}

\maketitle              

\begin{abstract}
Today's Large Language Models (LLMs) have showcased exemplary capabilities, ranging from simple text generation to advanced image processing. Such models are currently being explored for in-vehicle services such as supporting perception tasks in Advanced Driver Assistance Systems (ADAS) or Autonomous Driving (AD) systems, given the LLMs' capabilities to process multi-modal data. However, LLMs often generate nonsensical or unfaithful information, known as ``hallucinations'': a notable issue that needs to be mitigated. In this paper, we systematically explore the adoption of SelfCheckGPT to spot hallucinations by three state-of-the-art LLMs (GPT-4o, LLaVA, and Llama3) when analysing visual automotive data from two sources: Waymo Open Dataset, from the US, and PREPER CITY dataset, from Sweden. Our results show that GPT-4o is better at generating faithful image captions than LLaVA, whereas the former demonstrated leniency in mislabeling non-hallucinated content as hallucinations compared to the latter. Furthermore, the analysis of the performance metrics revealed that the dataset type (Waymo or PREPER CITY) did not significantly affect the quality of the captions or the effectiveness of hallucination detection. However, the models showed better performance rates over images captured during daytime, compared to during dawn, dusk or night. Overall, the results show that SelfCheckGPT and its adaptation can be used to filter hallucinations in generated traffic-related image captions for state-of-the-art LLMs.

\keywords{hallucination detection  \and safety-critical systems \and multi-modal data \and perception systems \and automotive \and large language models.}
\end{abstract}

\section{Introduction}

State-of-the-art Large Language Models (LLMs) have demonstrated remarkable capabilities in performing generative tasks. Nowadays, such generative tasks have progressed from simple text generation to advanced image generation involving multi-modal data. The usage of LLMs has been positively increased up to a level, where even standardized knowledge tests are already questioned~\cite{wang2024large}. Hence, LLMs such as Pre-trained Transformers (GPT) are adopted in many domains given their exceptional capabilities in language understanding and generation~\cite{generativeAI_progress}.

\subsection{Problem Domain and Motivation}

The proprietary LLMs such as GPT-4o introduced in May 2024~\cite{gpt4oWebsite} and open source models such as Large Language-and-Vision Assistant (LLaVA)~\cite{NEURIPS2023_6dcf277e} have been trained on a large corpus that contains text and image-based data. ome automotive Original Equipment Manufacturers (OEMs) are already experimenting with potential application scenarios where LLMS are used within vehicles to provide better services to their passengers by engaging in natural language-based conversations~\cite{bmwCarExpert,bmwWebsite}. As LLMs show great potential in image description tasks where the retrieved image captions are often well composed, it is not surprising that such LLMs could be even considered to improve perception systems for Advanced Driver Assistance Systems (ADAS) or Autonomous Driving (AD).

However, tackling the impact of LLM's stochasticity remains a challenge due to a notable issue known as hallucinations, which refers to the tendency of LLMs generating nonsensical information~\cite{huang2023survey_hallucinations}. Hallucinations caused by LLMs are unacceptable regardless of their usage scenario. Therefore many researchers have focused on hallucination detection and mitigation techniques~\cite{tonmoy2024comprehensivesurveyhallucinationmitigation} that depend on different approaches such as (a) combinations of retrieval-augmented generation (RAG)~\cite{es2023ragas,yu2023improving}, (b) comparing the generated response with the given ground truth~\cite{hartvigsen2024aging,guan2024mitigating}, (c) evaluating the LLM's own consistency in the generated responses~\cite{ronanki2022chatgpt}, or (d) systematically assessing whether excerpts of an LLM's generated answer can be substantiated with other responses obtained from it for the same prompt~\cite{manakul2023selfcheckgpt}.

\subsection{Research Goal and Research Questions}
\label{sec:researchQuestions}

Manakul et al.~\cite{manakul2023selfcheckgpt} have evaluated and extended a technique to spot hallucinations called SelfCheckGPT on a text corpus based on the information extracted from Wikibio dataset~\cite{lebret2016neuraltextgenerationstructured}. However, the application and adoption of SelfCheckGPT for usage scenarios covering multi-modal data such as images and text are currently the subject of ongoing research as outlined in Sec.~\ref{sec:relatedWork}. Hence, the goal of our research is to (a) adopt the SelfCheckGPT approach for multi-modal data from the automotive context that is relevant for ADAS and AD, and (b) to assess its performance across three state-of-the-art LLMs, namely GPT-4o, LLaVA, and Llama3, by using our datasets' labels as ground truth for reference. We derive the following research questions:

\begin{enumerate}[leftmargin=*, label={\textbf{RQ-\arabic*}}]
    \item To what extent can the SelfCheckGPT approach be adopted to spot potential hallucinations when using state-of-the-art LLMs for image captioning tasks for automotive usage scenarios?
    \item What is the performance of the adopted SelfCheckGPT approach on two state-of-the-art automotive datasets (Waymo covering traffic scenarios in the US, and PREPER CITY covering traffic scenarios in Sweden)?
    \item To what extent is the performance of SelfCheckGPT affected by environmental conditions such as light or weather?
\end{enumerate}

\subsection{Contributions and Scope}

We explore the adoption of SelfCheckGPT as the first study that aims at spotting potential hallucinations for automotive usage scenarios relevant for ADAS and AD. Our main contribution is the systematic performance evaluation of SelfCheckGPT and its adaptation to spot the hallucinations on two datasets from two geographical regions covering urban and suburban areas in the US and Sweden, normalized wrt.~the traffic scenarios covered in the respective datasets. Furthermore, the potential impact of the time of the day on the performance of SelfCheckGPT was assessed. We limited the captioning capabilities on vehicles, pedestrians, and cyclists for experimental reasons; allowing an LLM to freely describe everything it \emph{sees} in an image would maybe unveil more insights but would limit the scalability of the experimental setup. We propose an adaptation of SelfCheckGPT and its extension CrossCheckGPT~\cite{sun2024crosscheckgpt} to identify hallucinations in automotive usage scenarios.

\subsection{Structure of the Paper}
The remainder of the paper is organized as follows: Section \ref{sec:relatedWork} reviews existing hallucination detection and mitigation strategies. Section \ref{sec:methodology} provides the overview and details of our research methodology. Section \ref{sec:results} and Section \ref{sec:AnalysisAndDiscussion} present the results of the experiments and its analysis and discussion. Section~\ref{sec:conclusion} concludes the paper.

\section{Related Work}
\label{sec:relatedWork}

We reviewed adoptions and usage scenarios of SelfCheckGPT \cite{manakul2023selfcheckgpt}, which presents a self-correction hallucination detection mechanism for text-based data. Existing hallucination detection and mitigation strategies consider SelfCheckGPT as the baseline. 

Sun et al.~\cite{sun2024crosscheckgpt} present CrossCheckGPT that assesses the responses generated by a multi-modal LLM using the evidence responses that are generated by a different set of such LLMs. This method is slightly different from SelfCheckGPT, which assesses the consistency of the generated response using the same model. The proposed method has been validated for image-to-text data using the MHaluBench benchmark \cite{chen2024unified}, which contains 1143 image captioning data records. The said captions and the images are not focused on the automotive domain and, therefore, may not include labels relevant for perception-related tasks in the automotive discipline. Deng et al.~\cite{deng2024seeing} propose a hallucination mitigation technique that evaluates the LLM-generated responses against captions generated by a CLIP model. The CLIP Score has been used to evaluate the primary response and the candidate sentences. Elaraby et al.~\cite{elaraby2023halo} also present an adoption of SelfCheckGPT called HaloCheck that demonstrates better estimations of the severity of the hallucinations by using knowledge injection. This method requires fine-tuning the model with domain-specific knowledge to gain better performance and that limits the applicability of HaloCheck for LLMs in general.

The studies such as Hartvigsen et al.~\cite{hartvigsen2024aging}, Guan et al.~\cite{guan2024mitigating}, Es et al.~\cite{es2023ragas}, and Yu et al.~\cite{yu2023improving} propose different adaptations of SelfCheckGPT that use the concept of Retrieval Augmented Generation (RAG) by passing context to the LLM along with the question. Guan et al.~\cite{guan2024mitigating} use knowledge graphs created based on a selected dataset to retrieve the context related to the query. Even though the main research goal of \cite{hartvigsen2024aging} is not hallucination detection and mitigation, they propose an adaptation of SelfCheckGPT that requires correct sentences from Wikipedia to mitigate potential hallucinations. Similarly, \cite{es2023ragas} also uses a custom-made dataset called WikiEval that covers data retrieved from 50 Wikipedia pages to generate context for RAG. \cite{yu2023improving} proposes prompting the same LLM with the initial primary response together with the context taken from the retrieved documents to reduce hallucinations by refining the response. All aforementioned RAG-based SelfCheckGPT adaptations require additional information sources that are referred to as context, which is difficult to retrieve in the automotive domain especially related to perception-related tasks.

In addition to that, there are more recent studies conducted focusing on both factuality and consistency of the responses. Ji et al.~\cite{ji2023towards} propose a hallucination mitigation technique for question-and-answer systems, where multiple prompting is involved. Under this approach, the factuality of the initial primary response is assessed by a scorer and the response will be continuously refined until it reaches the threshold value. A similar approach will then be applied to assess the consistency of the response. Wu et al.~\cite{wu2024logical} also present a new technique to mitigate the hallucinations by understanding the logical consistency of the primary response. This method requires prompting the LLM twice with questions regarding the attributes and objects in the primary response. These actuality and consistency checking mechanisms demonstrate promising results focusing on text-based generic data. 

Cole et al.~\cite{cole2023selectively} address the ``ambiguous questions'' problem in the domain of LLMs, a very common issue that occurs in text-based processing applications. Even though the main goal of the study is tightly coupled with handling ambiguous questions, the proposed approach can be applied to mitigate hallucinations caused by LLMs. The authors have presented the idea of using another or the same LLM to validate the initial responses with boolean answers.

\section{Methodology}
\label{sec:methodology}

We aim to address the following research objectives with our study:
\begin{enumerate}
    \item Adopting SelfCheckGPT for multi-modal, automotive data,
    \item Designing an experimental setup that addresses the issue of determining the correctness of a sentence $s_i$ (cf.~aspect (a) mentioned before) that does not require additional data such as the dataset ground truth for image captioning tasks for automotive usage,
    \item Assessing the performance of different combinations of LLMs to effectively spot hallucinations, and
    \item Evaluating SelfCheckGPT's and the proposed adaptation's sensitivity to external influences such as light and weather conditions.
\end{enumerate}

\begin{figure}
    \centering
    \includegraphics[width=0.85\linewidth]{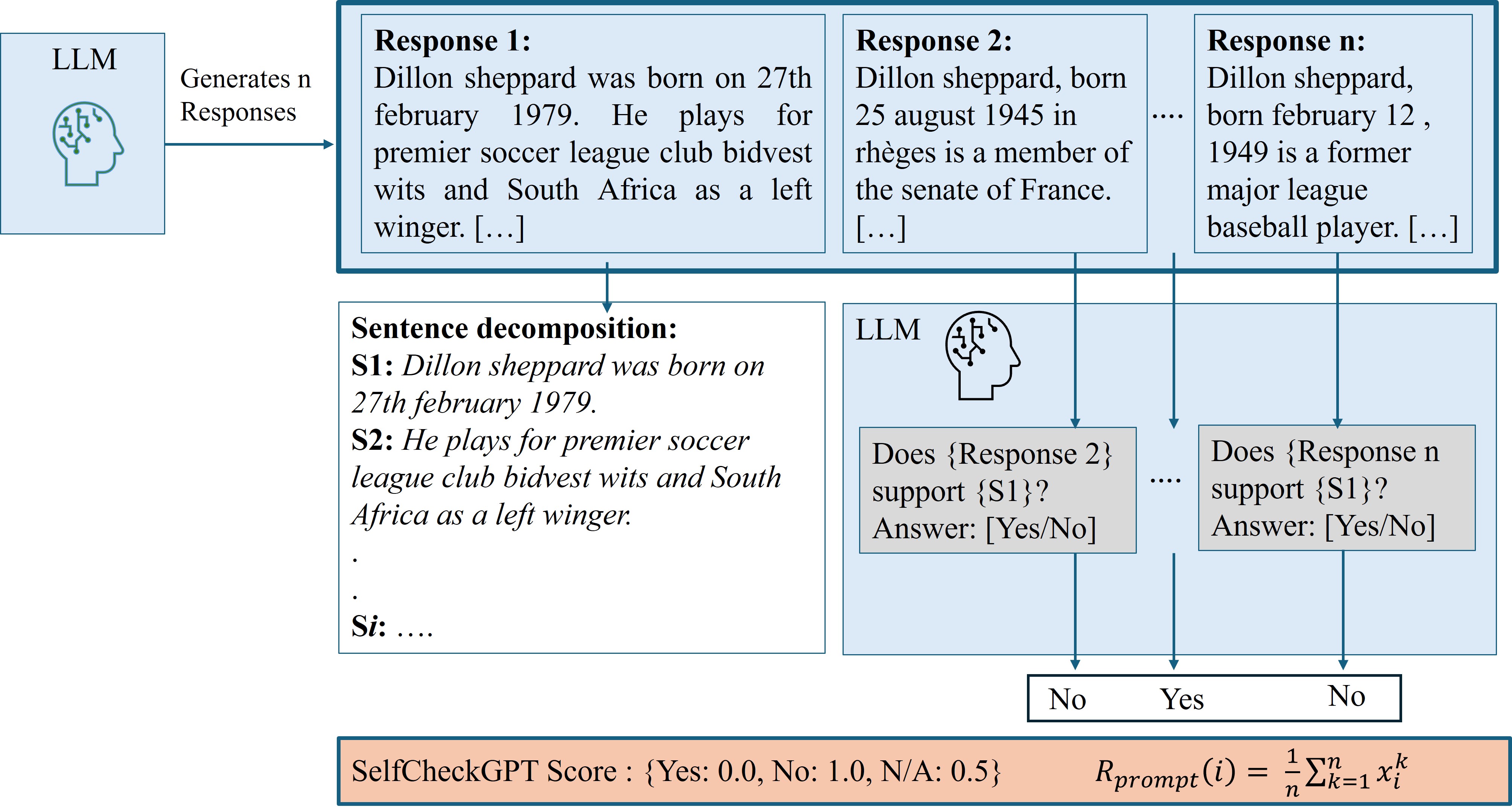}
    \caption{SelfCheckGPT with LLM prompting. The LLM-generated sentences in a caption are compared against the remaining captions generated by the same LLM for the same prompt. The sentences that are supported by the other captions are considered to be non-hallucinated and this comparison is conducted by LLMs.}
    \label{fig:selfcheck}
\end{figure}

As we adopt SelfCheckGPT for our setup, we describe its core principles in the following. The general idea behind SelfCheckGPT as depicted in Fig.~\ref{fig:selfcheck} is to sample a given LLM $n+1$ times for a specific prompt $P$. Then, the initial response $R_1$ provided by the LLM is divided into separate chunks of texts, for instance, separate sentences $s_1\cdots s_{n}$ from $R_1$. The consistency of SelfCheckGPT is measured using five variants including BERTScore, question-answering, n-gram, Natural Language Inference (NLI), and LLM prompting. We focus on the fifth variant ``SelfCheckGPT with LLM prompting'', given the effectiveness of LLMs in information assessing tasks \cite{guo2023evaluatinglargelanguagemodels}.  This variant uses an LLM to determine whether the subsequent responses $R_2, \cdots, R_{n+1}$ support the individual sentences $s_1 \cdots s_{n}$, respectively. For each $s_i$ from the initial response $R_1$, the same or a different LLM is prompted to check whether $s_i$ is supported by $R_{i+1}$. This is done by obtaining a \verb|yes| or \verb|no| reply for each check. The results from these individual consistency checks are aggregated to a joint score to spot potential hallucinations. The idea behind this is that either each sentence $s_i$ is not sufficiently supported by a $R_{i+1}$ or that a sufficiently large subset of responses is showing varying or contradicting support of the sentences. While this approach by design can neither provide proof of whether a given sentence $s_i$ is correct or incorrect nor show what part of a complete response is a hallucination, it yet allows to check for self-consistency and uses it as a proxy for detecting hallucinations. Assuming a certain level of internal consistency for the LLM in question, increasing the number of samples $n$ may enhance the likelihood of spotting potential hallucinations.

Our experiments consist of the following components: (A) Multi-modal, automotive datasets, and (B) LLMs that are capable of processing multi-modal prompts (ie., text and/or images simultaneously) as we presented selected images from different traffic situations with a task to the LLMs. To reduce specific, non-controllable, and potentially unknown influential factors of a given automotive dataset, we decided to use two different datasets: Waymo Open Dataset~\cite{waymoDataset} and PREPER CITY~\cite{revereDataset}. The Waymo Open Dataset was created in 2021 by Google in metropolitan areas in the US to support and facilitate research around algorithms needed for self-driving technology. The Waymo Open Dataset covers 2,030  segments, each approximately 20 seconds long. It contains around 390,000 captured video frames that cover five cameras including one forward-facing camera and four side cameras. 

As that dataset is US-centric and hence, specific to visual appearance of traffic agents like cars as well as driving styles typical to the US, numerous other datasets were created and shared over the years covering other regions of the world, featuring other sensors to capture a vehicle's surroundings, focusing other traffic situations. To complement the Waymo Open Dataset as well as to reduce its potential shortcomings, we included PREPER CITYwhich was collected in 2021 in Gothenburg, Sweden, and hence, covers other types of vehicles, metropolitan appearance, and different behavior in traffic from the included traffic actors. It features 114 traffic segments, each approximately 15 minutes long. It contains more than 1.5 million video frames covering multiple cameras. 

\subsection{Dataset Curation for Waymo and PREPER CITY}

Both datasets contain manually added annotations to foster the research and development of algorithms for ADAS and AD systems. These annotations are necessary to train, test, and evaluate the performance of specifically trained machine learning (ML) components to support a vehicle's perception stack. We use these labels (a) as ground truth to \emph{fact-check} the individual sentences $s_i$ from a generated response $R_j$ to assess the quality of initial answers from an LLM by comparing with the ground truth (for instance, if the labels state \verb|car| and \verb|truck|, but the LLM described \verb|car| and \verb|bike|; here, the LLM hallucinated the \verb|bike| and it also overlooked the \verb|truck|); furthermore, (b) we also used the ground truth to get an overview of the typical distribution of scenarios covered in the two datasets so that we sample similar traffic situations from both datasets; and finally, (c) we used the different label categories such as \verb|car|, \verb|truck|, \verb|pedestrian|, \verb|cyclist|, dots to consolidate a common super-set of keywords that we allowed the LLMs to use for its description.

The consolidation of keywords enabled the comparison of the generated responses with the ground truth for the two datasets. We heuristically determined a prompt for GPT-4o and LLaVA that allowed them to be as expressive as possible while constraining the description of traffic actors to be identified to match with our consolidated list of annotations that are valid on both datasets, which allowed to scale the number of different traffic situations in our experiments while relying on the ground truth labels for fact-checking.

Eventually, we conducted our experiments with the following curated subset of traffic scenarios: We selected 920 images from the Waymo Open Dataset, and another 920 images from PREPER CITY showing different combinations of traffic agents. 617 (PREPER CITY) and 619 (WAYMO) images contain only vehicles, whereas 10 (PREPER CITY) and 5 (WAYMO) images contain only pedestrians. 4 images from PREPER CITY contain only cyclists. Similarly, 165 (PREPER CITY) and 198 (WAYMO) images contain both vehicles and pedestrians whereas it is 31 (PREPER CITY) images for vehicles and cyclists and 6 (PREPER CITY) images for pedestrians and cyclists.  87 (PREPER CITY) and 98 (WAYMO) images contain all three traffic agents. The label \verb|vehicle| dominates the traffic scenarios captured in both datasets.

\subsection{Experimental Setup}

We depicted our experimental setup in Fig.~\ref{fig:Methodology}. For both multi-modal LLMs, GPT-4o (gpt-4o-2024-05-13 version) and LLaVA (latest 8dd30f6b0cb1 version), we fed every image 5 times using the following prompt as shown by step (A):

\begin{figure}
    \centering
    \includegraphics[width=1\linewidth,trim={0cm 0cm 0cm 0cm},clip]{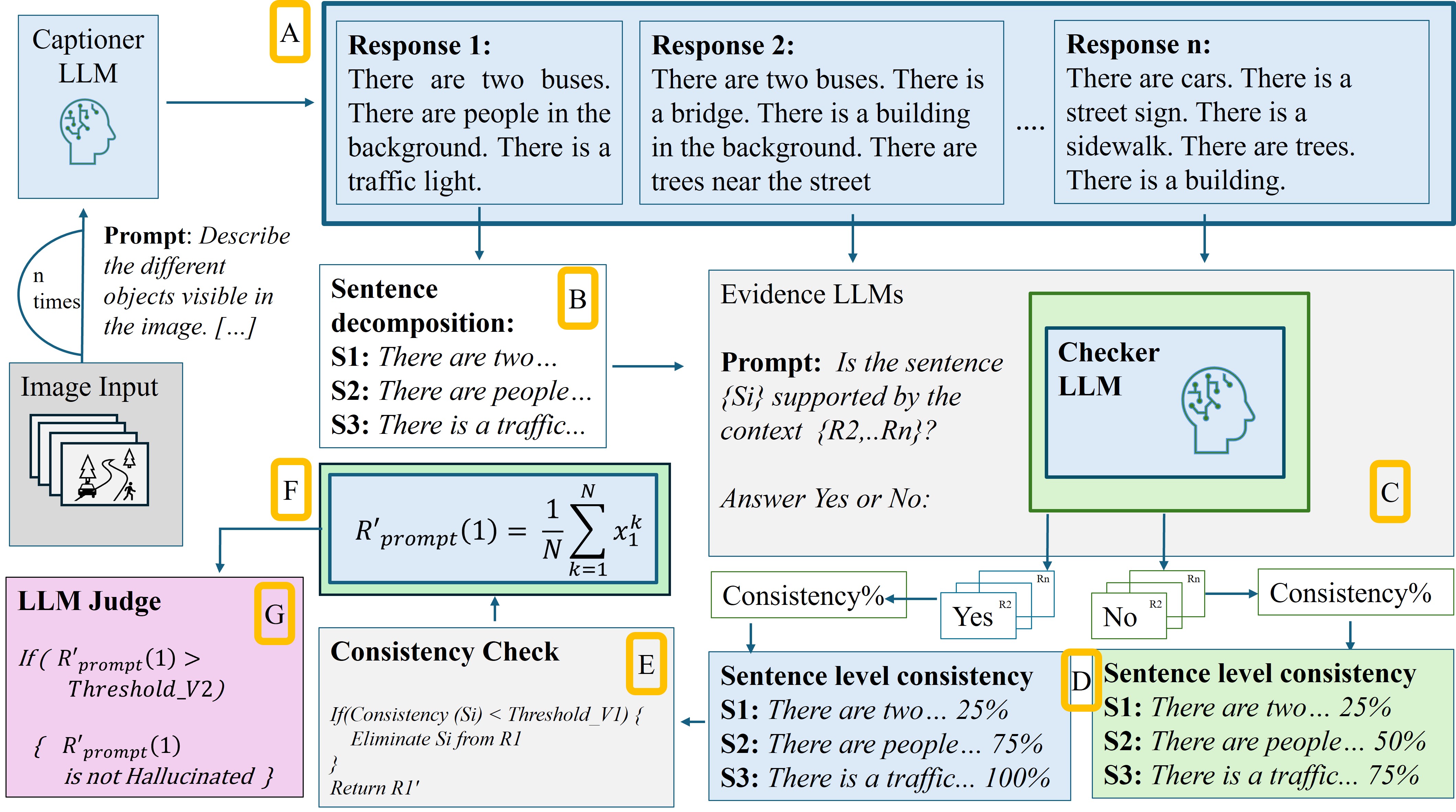}
    \caption{The experimental setup that depicts the adaptation of SelfCheckGPT. The LLM-generated sentences in a caption are compared with the remaining captions to identify the hallucinated sentences. Based on the sentence level consistency check, the sentences in the caption are filtered to create a refined version of the caption. Different checker and captioner LLMs are used in this setup.}
    \label{fig:Methodology}
\end{figure}

\begin{verbatim}
Describe the different objects visible in the image. Please write 
very simple and clear sentences. Use the format: "There are
[object]." For example, "There are cars. There are people. There
are cyclists."
Look carefully and make sure to mention all types of objects you
see, especially people. There are multiple types of objects in
the image, provide a separate sentence for each type.
\end{verbatim}

For each response $R_i$, we recorded the response itself for post-processing and the actual processing time per frame. Next, we broke down the first response $R_1$ into the individual sentences $s_{1 \dots n}$ as portrayed by step (B) in Fig.~\ref{fig:Methodology}. For each sentence $s_i$, we extracted the first noun/noun and determinant block and checked whether it matches the ground truth labels for that given image. This way, we could determine the sensitivity and specificity of an LLM's response: TP (non-hallucinations, not flagged as hallucinations), TN (hallucinations flagged as hallucinations), FP (hallucinations, not flagged as hallucinations), and FN (non-hallucinations, flagged as hallucinations).

Next, we applied the fifth variant of the SelfCheckGPT approach to determine for every sentence $s_i$ from $R_1$, whether it is supported by $R_{2 \dots n}$. This step is depicted by step (C) in the Fig.~\ref{fig:Methodology}. For each sentence $s_i$ and for each response $R_j$, we used the following prompt to obtain the \verb|Yes| or \verb|No| answer to calculate the potential hallucination score:

\begin{verbatim}
Context: {{CONTEXT}}  Sentence: {{SENTENCE}}
Is the sentence supported by the context above? Answer Yes or No:
\end{verbatim}

After calculating the sentence level consistency percentage for all sentences in $R_1$, the sentences with lower consistency levels were eliminated from the $R_1$, providing the opportunity to return a refined version of $R_1$ denoted by $R'_1$.These steps are showcased by steps (D) and (E) in Fig.~\ref{fig:Methodology}. In this experimental setup, we calculated the average consistency level for the caption by considering the average of the sentence level consistencies generated at step (D) based on the refined $R_1$. The caption level consistency percentage was used in step (G), where the LLM uses a threshold value to determine whether the refined $R_1$ is hallucinated or not. 

We also studied the performance of the hallucination score computations by combining permutations of different LLMs for the self-consistency checking. In any case, GPT-4o and LLaVA were used as origin for the responses $R_{1 \dots n}$, but applying the SelfCheckGPT approach was conducted in various permutations involving GPT-4o, LLaVA, and Llama3 : GPT-4o to check GPT-4o and LLaVA generated captions, and Llama3 to check GPT-4o and LLaVA generated captions. For the original use case scenario as motivated in our introduction, in particular, the combination LLaVA to feed Llama3 is of interest as it could be executed entirely offline, ie., with no access to a cloud back-end infrastructure, as well as the models are not proprietary in that sense that traceability concerning what model and which version is in use is possible in contrast to GPT-4o.

\section{Results}
\label{sec:results}

We report the results based on two categories as mentioned in the Sec.~\ref{sec:NoExtras} and in Sec.~\ref{sec:NoMissingAgents} considering different perspectives. Sec.~\ref{sec:NoExtras} focuses on the concept of hallucination detection and, therefore, defines the correctness as not having nonsensical traffic agents present in the answer compared to the input image. This method identifies the hallucinated traffic agents in the LLM-generated response, but may not include details about all traffic agents in the image. However, when it comes to the automotive domain, it is adamant that we learn about all traffic agents present in the area through the perception system to make the automated decision-making process more accurate. Therefore, not overlooking the traffic agents present in the area is crucial for perception-based tasks in ADAS/AD. 

Considering the importance of not overlooking traffic agents, we report results under Sec.~\ref{sec:NoMissingAgents} defining the correctness as not overlooking traffic agents in the caption compared to the input image. However, this approach does not apply to the sentence level consistency check as a single sentence may not contain information about all the objects. 

We conducted some further analysis to understand the impact made by each dataset on hallucination detection and how the time of the day impacted hallucination detection. Section~\ref{sec:DatasetEffect} and Section~\ref{sec:TimeofDayEffect}, respectively, contain results for the two categories and the two definitions of correctness mentioned above. Tab.~\ref{tab:NotMissingAgents_ExtraInfo_scenariosTable} shows a sequence of images taken from the PREPER CITY dataset together with sample captions to illustrate the two different definitions of correctness and the consistency check between the captions. 

\begin{table}
\caption{Example of captions, correctness checks, and consistency checks for a sequence.}
\label{tab:NotMissingAgents_ExtraInfo_scenariosTable}
\resizebox{\textwidth}{!}{%
\begin{tabular}{|p{2cm}|p{2cm}|p{2cm}|p{2cm}|p{2cm}|p{2cm}|p{2cm}|p{2cm}|p{2cm}|}
\hline
Timestamp & $t_0$ & $t_1$ & $t_2$ & $t_3$ & $t_4$ & $t_5$ & $t_6$ & $t_7$ \\ \hline
Images & \includegraphics[width=2cm]{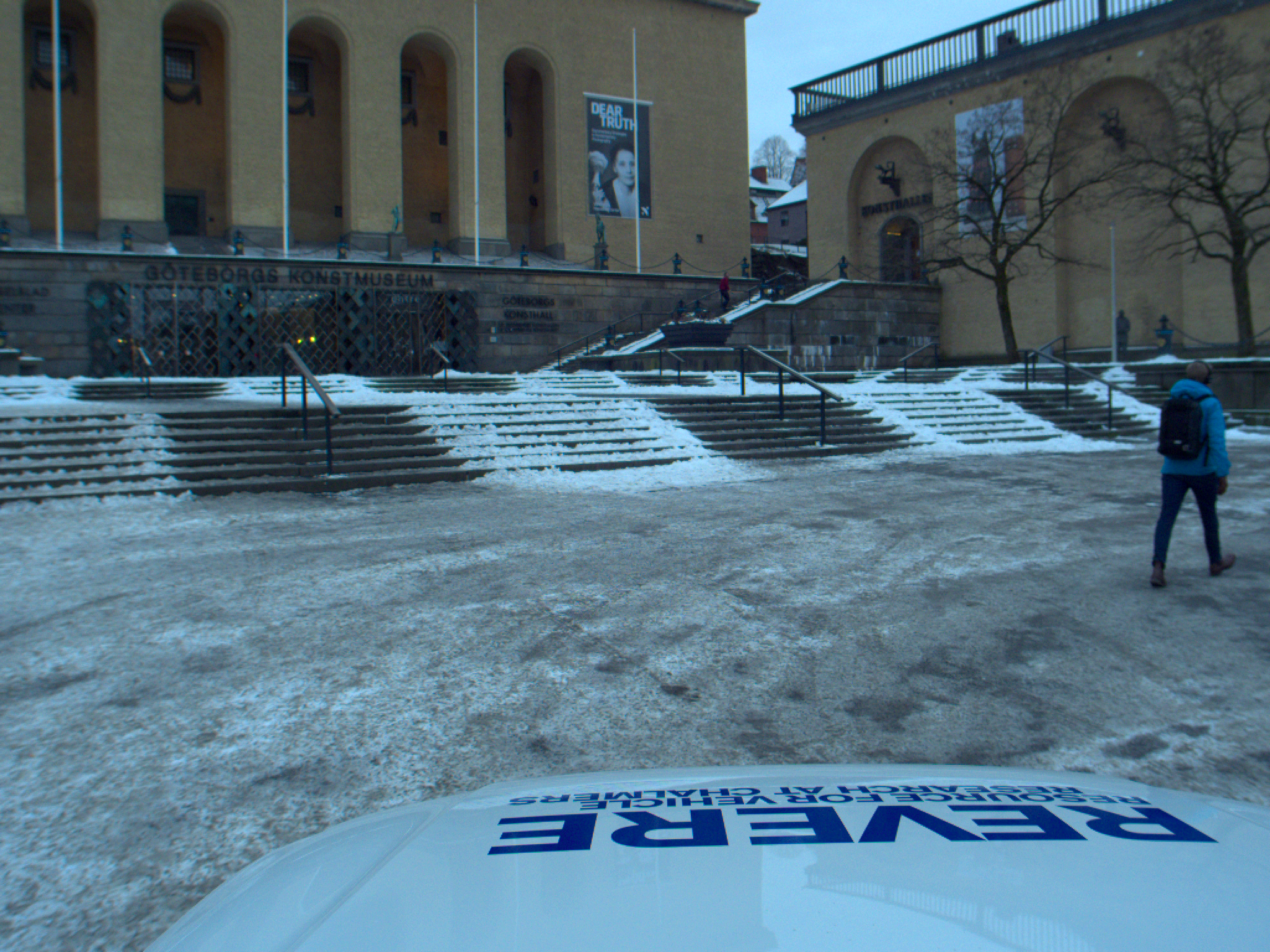} & \includegraphics[width=2cm]{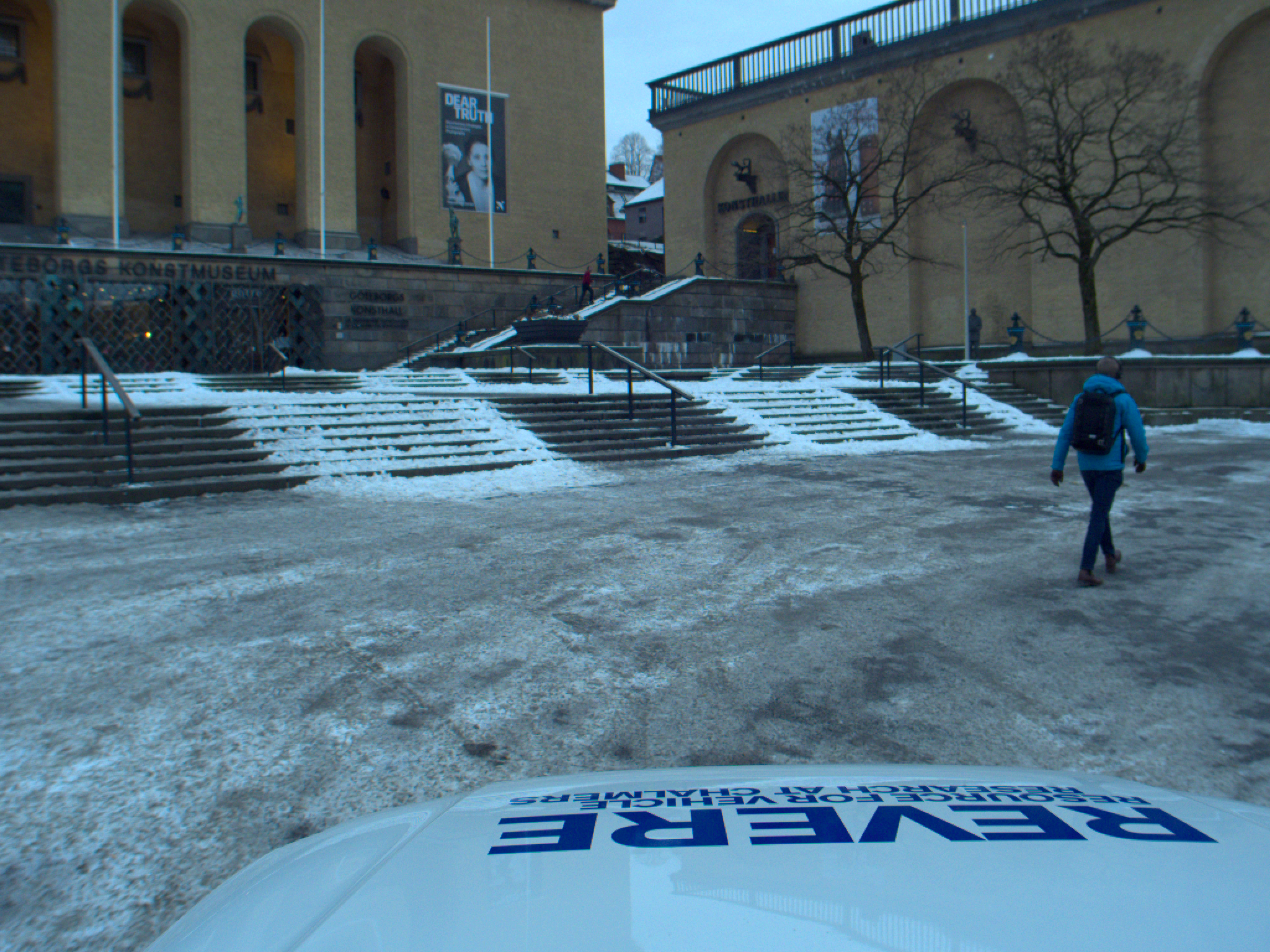} & \includegraphics[width=2cm]{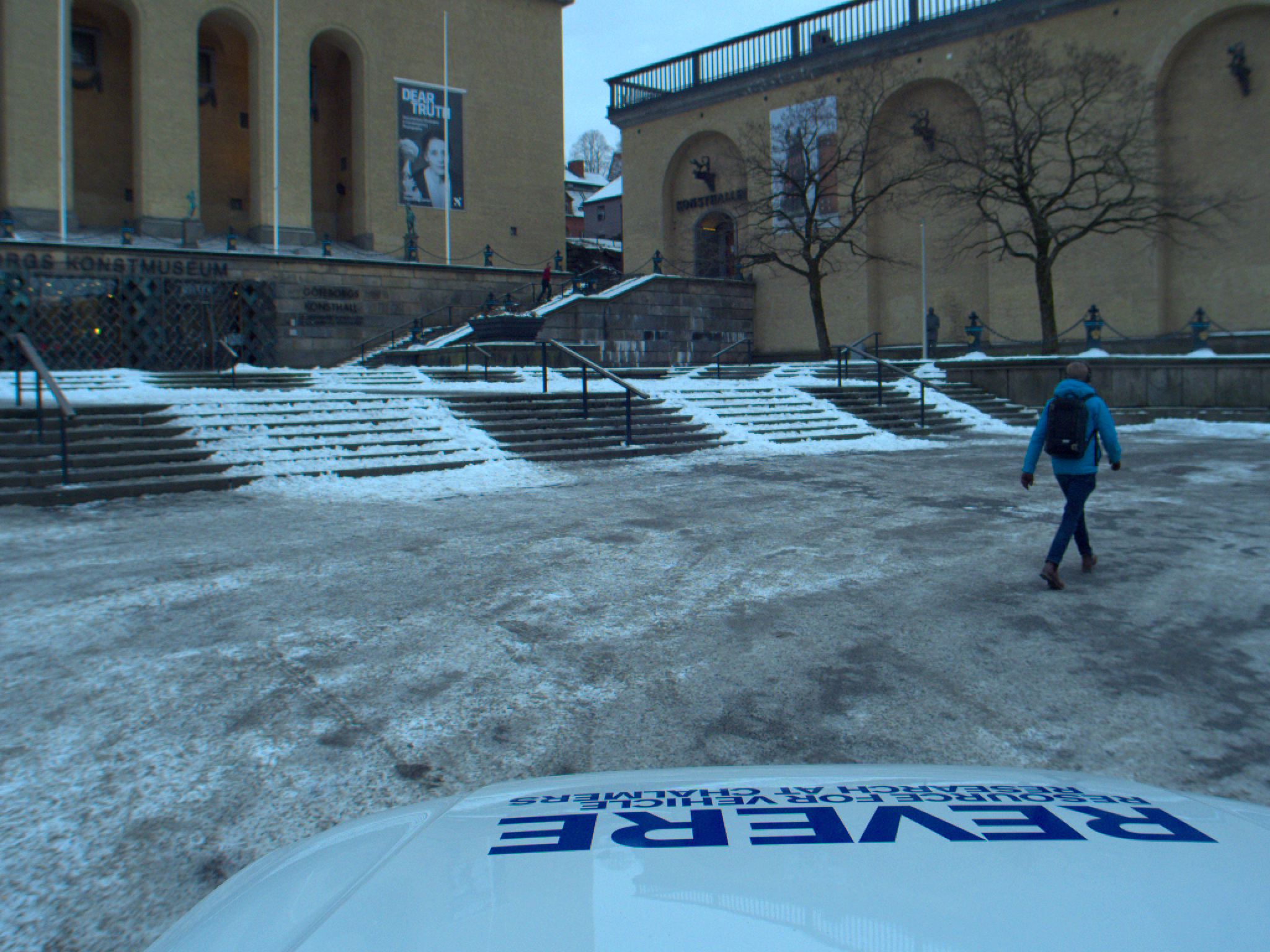} & \includegraphics[width=2cm]{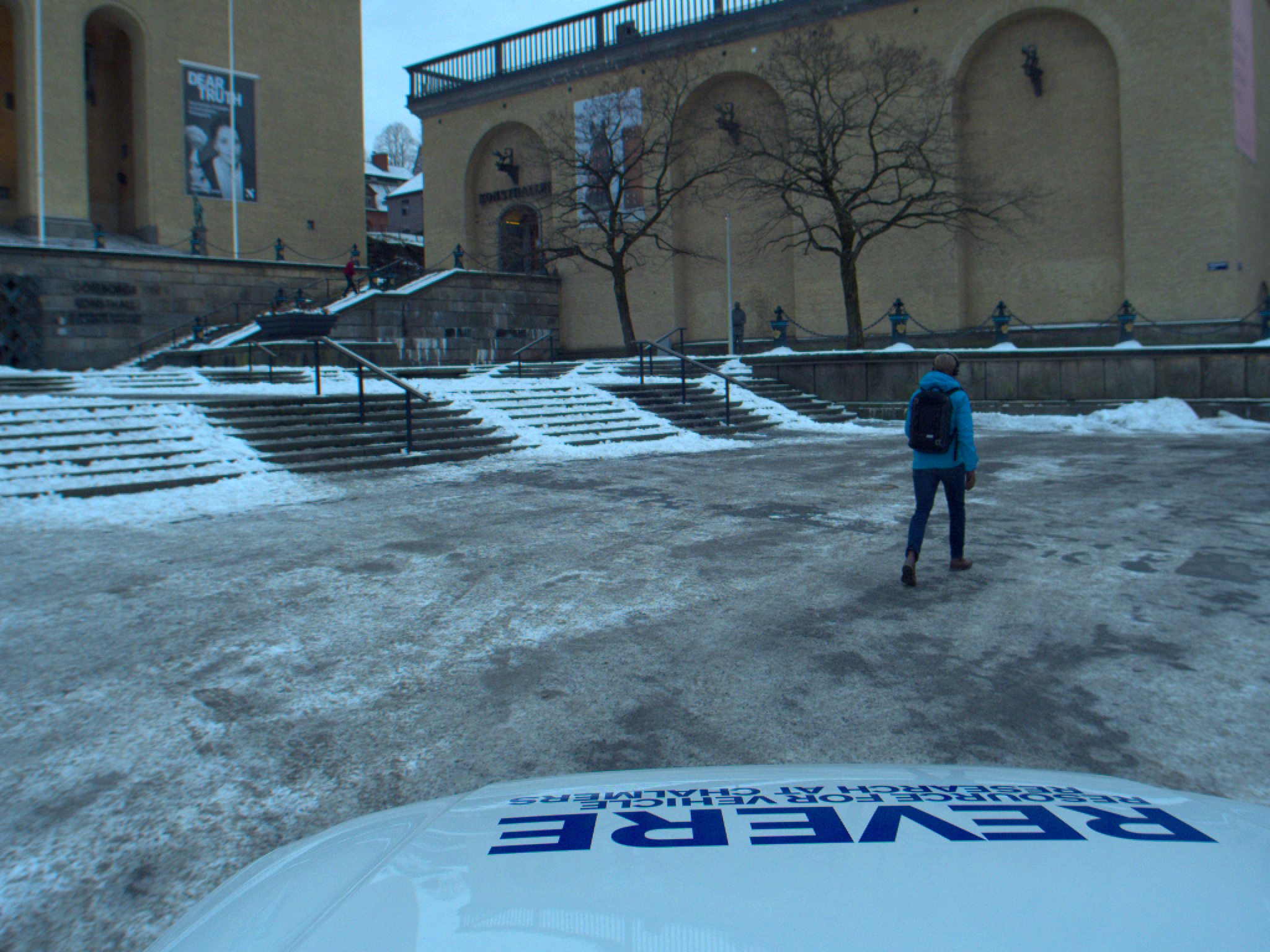} & \includegraphics[width=2cm]{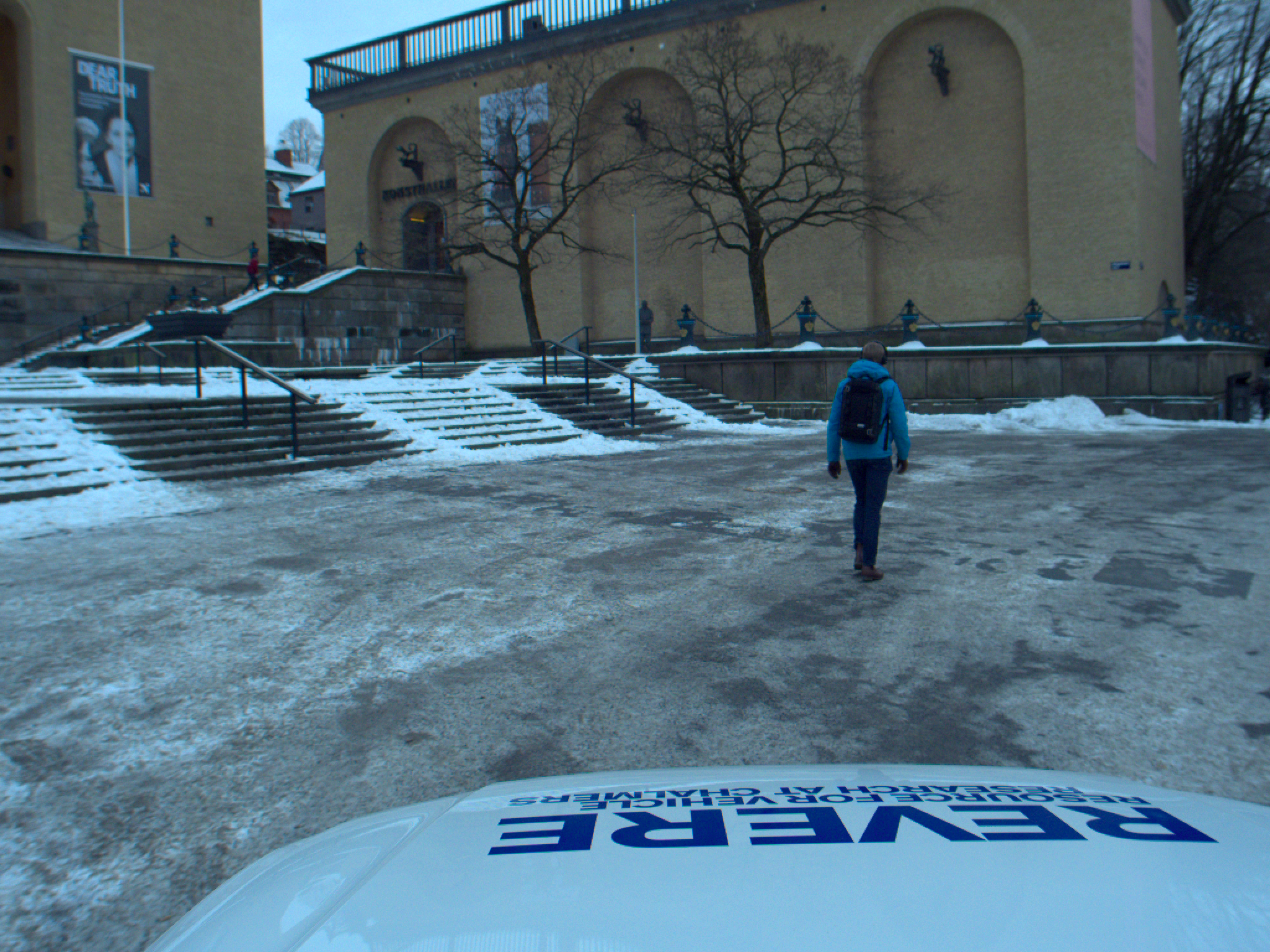} & \includegraphics[width=2cm]{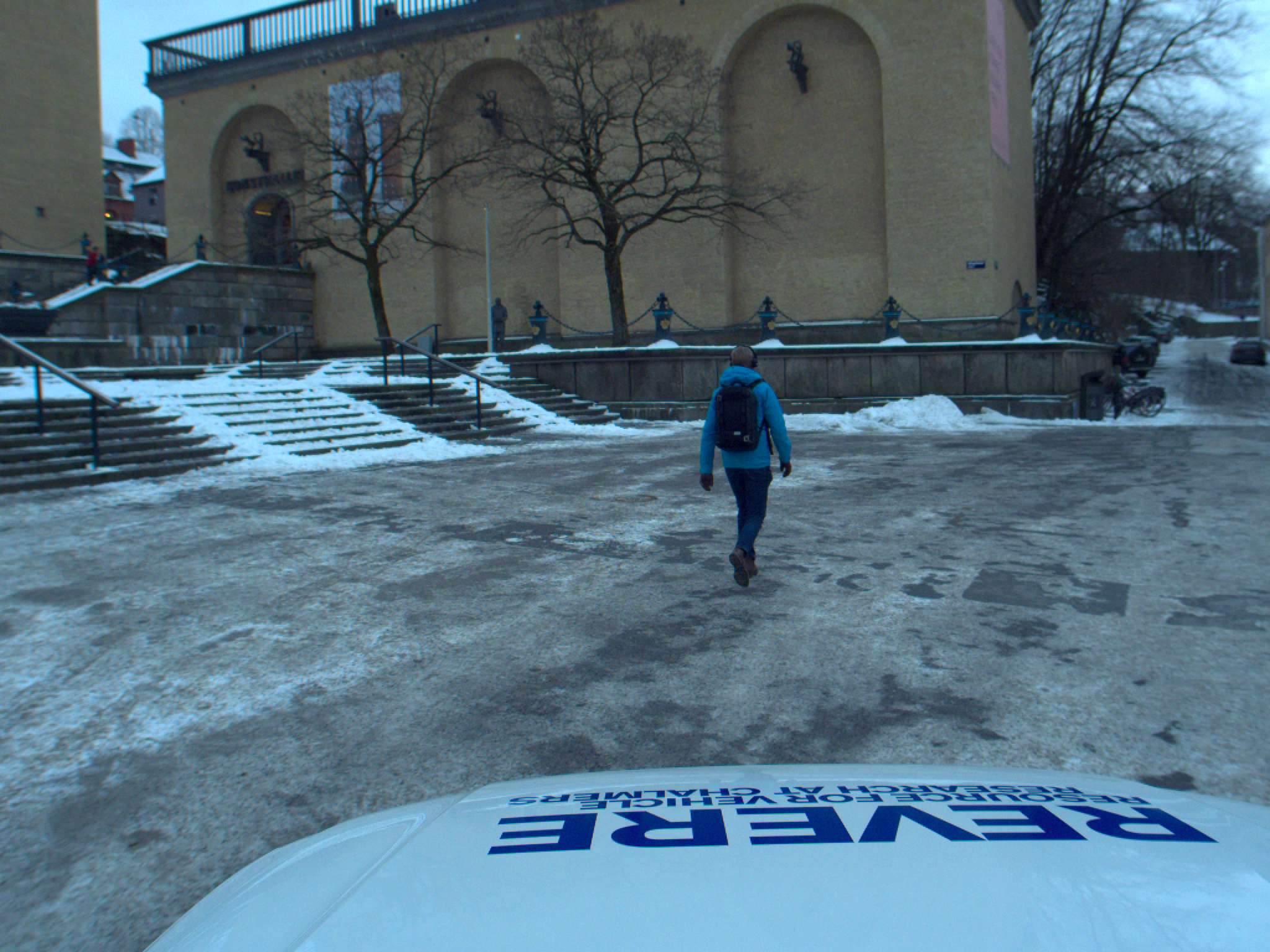} & \includegraphics[width=2cm]{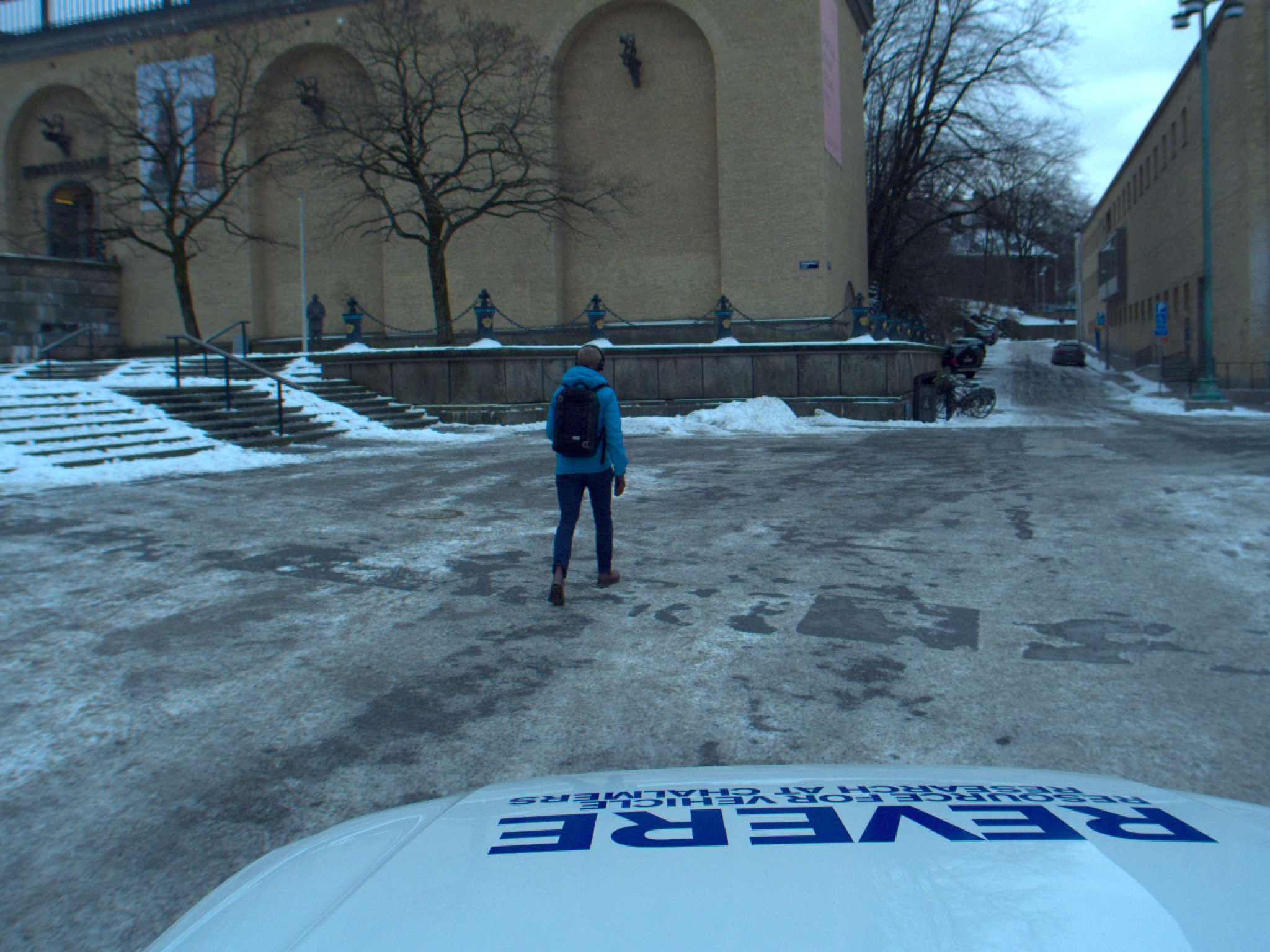} & \includegraphics[width=2cm]{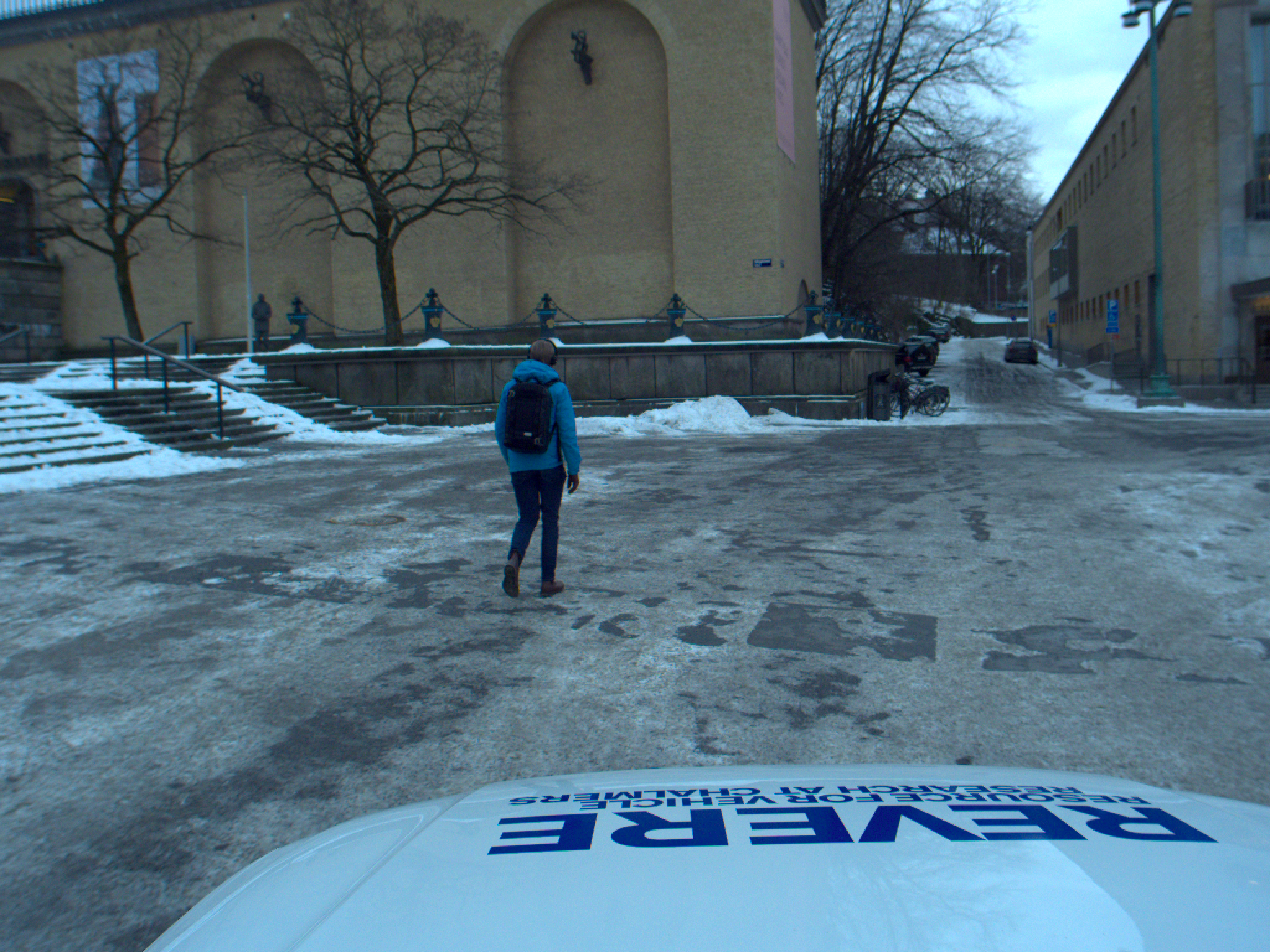} \\ \hline
Manual annotation & \{pedestrian\} & \{pedestrian\} & \{pedestrian\} & \{pedestrian\} & \{pedestrian\} & \{pedestrian\} & \{pedestrian\} & \{pedestrian\} \\ \hline
Caption 1 & There is a pedestrian & There is a pedestrian and a vehicle & There is a tree & There is a vehicle & There is a pedestrian & There is a pedestrian and a vehicle & There is a tree & There is a vehicle \\ \hline
Complimentary captions for check & There is a vehicle & There is a vehicle & There is a vehicle & There is a pedestrian & There is a pedestrian & There is a pedestrian and a vehicle & There is a tree & There is a vehicle \\ \hline
Captions consistent & \multicolumn{1}{c|}{\xmark} & \multicolumn{1}{c|}{\xmark} & \multicolumn{1}{c|}{\xmark} & \multicolumn{1}{c|}{\xmark} & \multicolumn{1}{c|}{\cmark} & \multicolumn{1}{c|}{\cmark} & \multicolumn{1}{c|}{\cmark} & \multicolumn{1}{c|}{\cmark} \\ \hline
Caption 1 - No hallucinations & \multicolumn{1}{c|}{\cmark} & \multicolumn{1}{c|}{\xmark} & \multicolumn{1}{c|}{\cmark} & \multicolumn{1}{c|}{\xmark} & \multicolumn{1}{c|}{\cmark} & \multicolumn{1}{c|}{\xmark} & \multicolumn{1}{c|}{\cmark} & \multicolumn{1}{c|}{\xmark} \\ \hline
Caption 1 - No overlooking & \multicolumn{1}{c|}{\cmark} & \multicolumn{1}{c|}{\cmark} & \multicolumn{1}{c|}{\xmark} & \multicolumn{1}{c|}{\xmark} & \multicolumn{1}{c|}{\cmark} & \multicolumn{1}{c|}{\cmark} & \multicolumn{1}{c|}{\xmark} & \multicolumn{1}{c|}{\xmark} \\ \hline
\end{tabular}%
}
\end{table}

The rest of the tables follow these definitions and show the performance metrics recorded for the original $R_1$ (the image caption before applying sentence filtering) and for the fixed response $R'_1$ (after applying sentence and caption filtering). For this, the sentence level consistency threshold has been arbitrarily fixed as 50\%. The metrics precision, recall, specificity, and F1 score help to understand the performance of the systems whereas the Matthews correlation coefficient helps to interpret more complex insights such as class imbalances and the performances of the models on minor classes. The results are recorded for both GPT-4o and Llama3 as checker LLMs considering the captions generated by LLaVA and GPT-4o as captioner LLMs.

\subsection{Detecting hallucinated traffic agents}
\label{sec:NoExtras}

The LLM-generated responses were checked against the ground truth annotations with the intention of hallucination detection as the baseline. Out of the captions generated by LLaVA and GPT-4o, 76.39\% and 94.51\%, respectively, were correct without any hallucinated content about the present traffic agents. These statistics are used as the baseline to evaluate the performances of the adapted methodology, which is reported in Tab.~\ref{tab:NOExtra_HallucinationDetection}.  

\begin{table}[]
\caption{Performance for hallucination detection using Llama3 and GPT-4o for the captions generated by GPT-4o and LLaVA. Caption correctness is defined as not hallucinating traffic agents. The performances are compared for the original response $R_1$, before filtering sentences, and for the fixed response, $R'_1$.}
\resizebox{\textwidth}{!}{%
\begin{tabular}{l|cccc|cccc|}
\cline{2-9}
                                                           & \multicolumn{4}{l|}{Fixed response $R'_1$}                                                                             & \multicolumn{4}{l|}{Original response $R_1$}                                                                           \\ \hline
\multicolumn{1}{|l|}{Captioner LLM}                        & \multicolumn{1}{c|}{LLaVA}   & \multicolumn{1}{c|}{\textbf{GPT4o}}   & \multicolumn{1}{c|}{LLaVA}   & \textbf{GPT4o}   & \multicolumn{1}{c|}{LLaVA}   & \multicolumn{1}{c|}{\textbf{GPT4o}}   & \multicolumn{1}{c|}{LLaVA}   & \textbf{GPT4o}   \\ \hline
\multicolumn{1}{|l|}{Checker LLM}                          & \multicolumn{1}{c|}{Llama3}   & \multicolumn{1}{c|}{\textbf{Llama3}}   & \multicolumn{1}{c|}{GPT4o}   & \textbf{GPT4o}   & \multicolumn{1}{c|}{Llama3}   & \multicolumn{1}{c|}{\textbf{Llama3}}   & \multicolumn{1}{c|}{GPT4o}   & \textbf{GPT4o}   \\ \hline
\multicolumn{1}{|l|}{Precision (correct over consistent)}  & \multicolumn{1}{c|}{86.23\%} & \multicolumn{1}{c|}{\textbf{96.38\%}} & \multicolumn{1}{c|}{92.89\%} & \textbf{96.92\%} & \multicolumn{1}{c|}{86.23\%} & \multicolumn{1}{c|}{\textbf{96.38\%}} & \multicolumn{1}{c|}{87.63\%} & \textbf{96.46\%} \\ \hline
\multicolumn{1}{|l|}{Recall (consistent over correct)}     & \multicolumn{1}{c|}{98.53\%} & \multicolumn{1}{c|}{\textbf{99.68\%}} & \multicolumn{1}{c|}{78.21\%} & \textbf{93.78\%} & \multicolumn{1}{c|}{98.53\%} & \multicolumn{1}{c|}{\textbf{99.68\%}} & \multicolumn{1}{c|}{80.1\%}  & \textbf{93.99\%} \\ \hline
\multicolumn{1}{|l|}{Specificity (flagged hallucinations)} & \multicolumn{1}{c|}{0.0\%}   & \multicolumn{1}{c|}{\textbf{0.0\%}}   & \multicolumn{1}{c|}{13.85\%} & \textbf{2.08\%}  & \multicolumn{1}{c|}{0.0\%}   & \multicolumn{1}{c|}{\textbf{0.0\%}}   & \multicolumn{1}{c|}{29.86\%} & \textbf{8.47\%}  \\ \hline
\multicolumn{1}{|l|}{F1 Score}                             & \multicolumn{1}{c|}{91.97\%} & \multicolumn{1}{c|}{\textbf{98.0\%}}  & \multicolumn{1}{c|}{84.92\%} & \textbf{95.32\%} & \multicolumn{1}{c|}{91.97\%} & \multicolumn{1}{c|}{\textbf{98.0\%}}  & \multicolumn{1}{c|}{83.7\%}  & \textbf{95.21\%} \\ \hline
\multicolumn{1}{|l|}{Matthews Correlation Coefficient}     & \multicolumn{1}{c|}{-0.0450} & \multicolumn{1}{c|}{\textbf{-0.0107}} & \multicolumn{1}{c|}{-0.0478} & \textbf{-0.0292} & \multicolumn{1}{c|}{-0.0450} & \multicolumn{1}{c|}{\textbf{-0.0107}} & \multicolumn{1}{c|}{0.0841}  & \textbf{0.0193}  \\ \hline
\end{tabular}
}
\label{tab:NOExtra_HallucinationDetection}
\end{table}

\subsection{Trusting captions: Detecting overlooked traffic agents}
\label{sec:NoMissingAgents}

Considering the correctness based on not overlooking traffic agents, the LLaVA and GPT4o-generated captions were again checked against the ground truth annotations to use as a baseline. 75.1\% of the captions in the GPT4o-generated captions were reported as correct whereas the correct percentage was 76.95\%  for the LLaVA-generated captions. Tab.~\ref{tab:NotOverlookingAgents} contains the performance metrics for the adapted methodology ``Not overlooking traffic agents'' that can be compared against the above correct percentages as a baseline. 

\begin{table}[]
\caption{Performance of spotting overlooking traffic agents using Llama3 and GPT-4o in the captions generated by GPT-4o and LLaVA. Correctness is defined as not overlooking traffic agents in the captions. The performances are compared for the original response $R_1$, before filtering sentences, and for the fixed response $R'_1$.}
\resizebox{\textwidth}{!}{%
\begin{tabular}{l|cccc|cccc|}
\cline{2-9}
                                                           & \multicolumn{4}{l|}{Fixed response $R'_1$}                                                                             & \multicolumn{4}{l|}{Original response $R_1$}                                                                           \\ \hline
\multicolumn{1}{|l|}{Captioner LLM}                        & \multicolumn{1}{c|}{LLaVA}   & \multicolumn{1}{c|}{\textbf{GPT4o}}   & \multicolumn{1}{c|}{LLaVA}   & \textbf{GPT4o}   & \multicolumn{1}{c|}{LLaVA}   & \multicolumn{1}{c|}{\textbf{GPT4o}}   & \multicolumn{1}{c|}{LLaVA}   & \textbf{GPT4o}   \\ \hline
\multicolumn{1}{|l|}{Checker LLM}                          & \multicolumn{1}{c|}{Llama3}   & \multicolumn{1}{c|}{\textbf{Llama3}}   & \multicolumn{1}{c|}{GPT4o}   & \textbf{GPT4o}   & \multicolumn{1}{c|}{Llama3}   & \multicolumn{1}{c|}{\textbf{Llama3}}   & \multicolumn{1}{c|}{GPT4o}   & \textbf{GPT4o}   \\ \hline
\multicolumn{1}{|l|}{Precision (correct over consistent)}  & \multicolumn{1}{c|}{72.78\%} & \multicolumn{1}{c|}{\textbf{73.59\%}} & \multicolumn{1}{c|}{72.65\%} & \textbf{72.72\%} & \multicolumn{1}{c|}{72.78\%} & \multicolumn{1}{c|}{\textbf{73.59\%}} & \multicolumn{1}{c|}{72.65\%} & \textbf{72.72\%} \\ \hline
\multicolumn{1}{|l|}{Recall (consistent over correct)}     & \multicolumn{1}{c|}{99.47\%} & \multicolumn{1}{c|}{\textbf{100.0\%}} & \multicolumn{1}{c|}{87.81\%} & \textbf{97.71\%} & \multicolumn{1}{c|}{98.43\%} & \multicolumn{1}{c|}{\textbf{99.75\%}} & \multicolumn{1}{c|}{78.1\%}  & \textbf{93.82\%} \\ \hline
\multicolumn{1}{|l|}{Specificity (flagged hallucinations)} & \multicolumn{1}{c|}{3.2\%}   & \multicolumn{1}{c|}{\textbf{1.15\%}}  & \multicolumn{1}{c|}{38.25\%} & \textbf{14.93\%} & \multicolumn{1}{c|}{0.47\%}  & \multicolumn{1}{c|}{\textbf{0.46\%}}  & \multicolumn{1}{c|}{19.59\%} & \textbf{5.88\%}  \\ \hline
\multicolumn{1}{|l|}{F1 Score}                             & \multicolumn{1}{c|}{84.06\%} & \multicolumn{1}{c|}{\textbf{84.78\%}} & \multicolumn{1}{c|}{79.51\%} & \textbf{83.38\%} & \multicolumn{1}{c|}{83.68\%} & \multicolumn{1}{c|}{\textbf{84.69\%}} & \multicolumn{1}{c|}{75.28\%} & \textbf{81.94\%} \\ \hline
\multicolumn{1}{|l|}{Matthews Correlation Coefficient}     & \multicolumn{1}{c|}{0.1069}  & \multicolumn{1}{c|}{\textbf{0.0919}}  & \multicolumn{1}{c|}{0.3034}  & \textbf{0.2423}  & \multicolumn{1}{c|}{-0.0434} & \multicolumn{1}{c|}{\textbf{0.0170}}  & \multicolumn{1}{c|}{-0.0249} & \textbf{-0.0054} \\ \hline

\end{tabular}
}
\label{tab:NotOverlookingAgents}
\end{table}

\subsection{Dataset effect on hallucination detection}
\label{sec:DatasetEffect}

This section presents the results based on each dataset considering the two approaches to understand a dataset's impact on the hallucination detection process in perception tasks targeting ADAS and AD. 
Firstly, we define correct captions as those not containing any traffic agents that are not mentioned by the manual labels.
Tab.~\ref{tab:Waymo_NO_Hallucinations} includes the performance metrics recorded for the images retrieved from the Waymo dataset following the correctness definition of ``detecting Hallucinations''. Tab.~\ref{tab:PREPER_NO_Hallucinations} includes the performance metrics recorded for the PREPER CITY dataset images. 
\begin{table}[]
\caption{Performance of Hallucination detection using Llama3 and GPT-4o for the captions generated by GPT-4o and LLaVA for Waymo images. Correctness is defined as not hallucinating traffic agents. The performances are compared for the original response $R_1$, before filtering sentences, and for the fixed response $R'_1$.}
\resizebox{\textwidth}{!}{%
\begin{tabular}{|l|cccc|cccc|}
\hline
\textbf{Dataset: Waymo}              & \multicolumn{4}{l|}{Fixed response $R'_1$}                                                                                 & \multicolumn{4}{l|}{Original response $R_1$}                                                                            \\ \hline
Captioner LLM                        & \multicolumn{1}{c|}{LLaVA}    & \multicolumn{1}{c|}{\textbf{GPT4o}}    & \multicolumn{1}{c|}{LLaVA}    & \textbf{GPT4o}    & \multicolumn{1}{c|}{LLaVA}    & \multicolumn{1}{c|}{\textbf{GPT4o}}   & \multicolumn{1}{c|}{LLaVA}   & \textbf{GPT4o}   \\ \hline
Checker LLM                          & \multicolumn{1}{c|}{Llama3}    & \multicolumn{1}{c|}{\textbf{Llama3}}    & \multicolumn{1}{c|}{GPT4o}    & \textbf{GPT4o}    & \multicolumn{1}{c|}{Llama3}    & \multicolumn{1}{c|}{\textbf{Llama3}}   & \multicolumn{1}{c|}{GPT4o}   & \textbf{GPT4o}   \\ \hline
Precision (correct over consistent)  & \multicolumn{1}{c|}{84.51\%}  & \multicolumn{1}{c|}{\textbf{96.8\%}}   & \multicolumn{1}{c|}{92.64\%}  & \textbf{97.31\%}  & \multicolumn{1}{c|}{84.38\%}  & \multicolumn{1}{c|}{\textbf{96.8\%}}  & \multicolumn{1}{c|}{86.95\%} & \textbf{96.93\%} \\ \hline
Recall (consistent over correct)     & \multicolumn{1}{c|}{99.2\%}   & \multicolumn{1}{c|}{\textbf{99.75\%}}  & \multicolumn{1}{c|}{81.2\%}   & \textbf{95.0\%}   & \multicolumn{1}{c|}{99.2\%}   & \multicolumn{1}{c|}{\textbf{99.75\%}} & \multicolumn{1}{c|}{84.12\%} & \textbf{95.22\%} \\ \hline
F1 Score                             & \multicolumn{1}{c|}{91.27\%}  & \multicolumn{1}{c|}{\textbf{98.25\%}}  & \multicolumn{1}{c|}{86.54\%}  & \textbf{96.14\%}  & \multicolumn{1}{c|}{91.19\%}  & \multicolumn{1}{c|}{\textbf{98.25\%}} & \multicolumn{1}{c|}{85.51\%} & \textbf{96.07\%} \\ \hline
Specificity (flagged hallucinations) & \multicolumn{1}{c|}{0.0\%}    & \multicolumn{1}{c|}{\textbf{0.0\%}}    & \multicolumn{1}{c|}{10.94\%}  & \textbf{0.0\%}    & \multicolumn{1}{c|}{0.0\%}    & \multicolumn{1}{c|}{\textbf{0.0\%}}   & \multicolumn{1}{c|}{31.29\%} & \textbf{7.69\%}  \\ \hline
Matthews Correlation Coefficient     & \multicolumn{1}{c|}{-0.03514} & \multicolumn{1}{c|}{\textbf{-0.00900}} & \multicolumn{1}{c|}{-0.05107} & \textbf{-0.03666} & \multicolumn{1}{c|}{-0.03532} & \multicolumn{1}{c|}{\textbf{-0.0090}} & \multicolumn{1}{c|}{0.14448} & \textbf{0.02369} \\ \hline

\end{tabular}
}
\label{tab:Waymo_NO_Hallucinations}
\end{table}
\begin{table}[]
\caption{Performance of Hallucination detection using Llama3 and GPT-4o for the captions generated by GPT-4o and LLaVA for PREPER CITY images. Correctness is defined as not hallucinating traffic agents. The performances are compared for the original response $R_1$, before filtering sentences, and for the fixed response $R'_1$.}
\resizebox{\textwidth}{!}{%
\begin{tabular}{|l|cccc|cccc|}
\hline
\textbf{Dataset: PREPER CITY}             & \multicolumn{4}{l|}{Fixed response $R'_1$}                                                                             & \multicolumn{4}{l|}{Original response $R_1$}                                                                           \\ \hline
Captioner LLM                        & \multicolumn{1}{c|}{LLaVA}   & \multicolumn{1}{c|}{\textbf{GPT4o}}   & \multicolumn{1}{c|}{LLaVA}   & \textbf{GPT4o}   & \multicolumn{1}{c|}{LLaVA}   & \multicolumn{1}{c|}{\textbf{GPT4o}}   & \multicolumn{1}{c|}{LLaVA}   & \textbf{GPT4o}   \\ \hline
Checker LLM                          & \multicolumn{1}{c|}{Llama3}   & \multicolumn{1}{c|}{\textbf{Llama3}}   & \multicolumn{1}{c|}{GPT4o}   & \textbf{GPT4o}   & \multicolumn{1}{c|}{Llama3}   & \multicolumn{1}{c|}{\textbf{Llama3}}   & \multicolumn{1}{c|}{GPT4o}   & \textbf{GPT4o}   \\ \hline
Precision (correct over consistent)  & \multicolumn{1}{c|}{87.78\%} & \multicolumn{1}{c|}{\textbf{95.95\%}} & \multicolumn{1}{c|}{93.14\%} & \textbf{96.51\%} & \multicolumn{1}{c|}{87.65\%} & \multicolumn{1}{c|}{\textbf{95.95\%}} & \multicolumn{1}{c|}{88.28\%} & \textbf{95.97\%} \\ \hline
Recall (consistent over correct)     & \multicolumn{1}{c|}{97.95\%} & \multicolumn{1}{c|}{\textbf{99.62\%}} & \multicolumn{1}{c|}{75.53\%} & \textbf{92.53\%} & \multicolumn{1}{c|}{98.22\%} & \multicolumn{1}{c|}{\textbf{99.62\%}} & \multicolumn{1}{c|}{76.62\%} & \textbf{92.73\%} \\ \hline
F1 Score                             & \multicolumn{1}{c|}{92.59\%} & \multicolumn{1}{c|}{\textbf{97.75\%}} & \multicolumn{1}{c|}{83.42\%} & \textbf{94.47\%} & \multicolumn{1}{c|}{92.64\%} & \multicolumn{1}{c|}{\textbf{97.75\%}} & \multicolumn{1}{c|}{82.04\%} & \textbf{94.32\%} \\ \hline
Specificity (flagged hallucinations) & \multicolumn{1}{c|}{0.0\%}   & \multicolumn{1}{c|}{\textbf{0.0\%}}   & \multicolumn{1}{c|}{16.67\%} & \textbf{3.7\%}   & \multicolumn{1}{c|}{1.94\%}  & \multicolumn{1}{c|}{\textbf{0.0\%}}   & \multicolumn{1}{c|}{28.24\%} & \textbf{9.09\%}  \\ \hline
Matthews Correlation Coefficient     & \multicolumn{1}{c|}{-0.0500} & \multicolumn{1}{c|}{\textbf{-0.0124}} & \multicolumn{1}{c|}{-0.0442} & \textbf{-0.0260} & \multicolumn{1}{c|}{0.0039}  & \multicolumn{1}{c|}{\textbf{-0.0124}} & \multicolumn{1}{c|}{0.0375}  & \textbf{0.0138}  \\ \hline

\end{tabular}
}
\label{tab:PREPER_NO_Hallucinations}
\end{table}

Secondly, we consider captions to be correct if they do not overlook any traffic agents that appear in the manual annotations.
Tab.~\ref{tab:Waymo_NotOverlookingAgents} includes the performance metrics recorded for the Waymo images following the correctness definition of ``not overlooking traffic agents''. 
\begin{table}[]
\caption{Performance of spotting overlooking traffic agents using Llama3 and GPT-4o in the captions generated by GPT-4o and LLaVA using Waymo images. Correctness is defined as not overlooking traffic agents in the captions. The performances are compared for the original response $R_1$, before filtering sentences, and for the fixed response $R'_1$.}
\resizebox{\textwidth}{!}{%
\begin{tabular}{|l|cccc|cccc|}
\hline
\textbf{Dataset: Waymo}              & \multicolumn{4}{l|}{Fixed response $R'_1$}                                                                             & \multicolumn{4}{l|}{Original response $R_1$}                                                                           \\ \hline
Captioner LLM                        & \multicolumn{1}{c|}{LLaVA}   & \multicolumn{1}{c|}{\textbf{GPT4o}}   & \multicolumn{1}{c|}{LLaVA}   & \textbf{GPT4o}   & \multicolumn{1}{c|}{LLaVA}   & \multicolumn{1}{c|}{\textbf{GPT4o}}   & \multicolumn{1}{c|}{LLaVA}   & \textbf{GPT4o}   \\ \hline
Checker LLM                          & \multicolumn{1}{c|}{Llama3}   & \multicolumn{1}{c|}{\textbf{Llama3}}   & \multicolumn{1}{c|}{GPT4o}   & \textbf{GPT4o}   & \multicolumn{1}{c|}{Llama3}   & \multicolumn{1}{c|}{\textbf{Llama3}}   & \multicolumn{1}{c|}{GPT4o}   & \textbf{GPT4o}   \\ \hline
Precision (correct over consistent)  & \multicolumn{1}{c|}{71.47\%} & \multicolumn{1}{c|}{\textbf{71.83\%}} & \multicolumn{1}{c|}{68.6\%}  & \textbf{68.37\%} & \multicolumn{1}{c|}{71.47\%} & \multicolumn{1}{c|}{\textbf{71.83\%}} & \multicolumn{1}{c|}{68.6\%}  & \textbf{68.37\%} \\ \hline
Recall (consistent over correct)     & \multicolumn{1}{c|}{99.62\%} & \multicolumn{1}{c|}{\textbf{100.0\%}} & \multicolumn{1}{c|}{88.5\%}  & \textbf{96.74\%} & \multicolumn{1}{c|}{99.06\%} & \multicolumn{1}{c|}{\textbf{99.66\%}} & \multicolumn{1}{c|}{79.73\%} & \textbf{93.85\%} \\ \hline
F1 Score                             & \multicolumn{1}{c|}{83.23\%} & \multicolumn{1}{c|}{\textbf{83.61\%}} & \multicolumn{1}{c|}{77.29\%} & \textbf{80.12\%} & \multicolumn{1}{c|}{83.03\%} & \multicolumn{1}{c|}{\textbf{83.49\%}} & \multicolumn{1}{c|}{73.75\%} & \textbf{79.11\%} \\ \hline
Specificity (flagged hallucinations) & \multicolumn{1}{c|}{1.41\%}  & \multicolumn{1}{c|}{\textbf{0.87\%}}  & \multicolumn{1}{c|}{29.97\%} & \textbf{8.18\%}  & \multicolumn{1}{c|}{0.0\%}   & \multicolumn{1}{c|}{\textbf{0.0\%}}   & \multicolumn{1}{c|}{13.52\%} & \textbf{1.98\%}  \\ \hline
Matthews Correlation Coefficient     & \multicolumn{1}{c|}{0.05692} & \multicolumn{1}{c|}{\textbf{0.07886}} & \multicolumn{1}{c|}{0.2303}  & \textbf{0.1072}  & \multicolumn{1}{c|}{-0.0518} & \multicolumn{1}{c|}{\textbf{-0.0310}} & \multicolumn{1}{c|}{-0.0797} & \textbf{-0.0892} \\ \hline

\end{tabular}
}
\label{tab:Waymo_NotOverlookingAgents}
\end{table}
Finally, Tab.~\ref{tab:PREPER_NotOverlookingAgents} includes the performance metrics recorded for the images retrieved from the PREPER CITY dataset following the correctness definition of ``Not overlooking traffic agents''.
\begin{table}[]
\caption{Performance of spotting overlooking traffic agents using Llama3 and GPT-4o in the captions generated by GPT-4o and LLaVA using PREPER CITY images. Correctness is defined as not overlooking traffic agents in the captions. The performances are compared for the original response $R_1$, before filtering sentences, and for the fixed response $R'_1$.}
\resizebox{\textwidth}{!}{%
\begin{tabular}{|l|cccc|cccc|}
\hline
\textbf{Dataset: PREPER CITY}             & \multicolumn{4}{l|}{Fixed response $R'_1$}                                                                             & \multicolumn{4}{l|}{Original response $R_1$}                                                                           \\ \hline
Captioner LLM                        & \multicolumn{1}{c|}{LLaVA}   & \multicolumn{1}{c|}{\textbf{GPT4o}}   & \multicolumn{1}{c|}{LLaVA}   & \textbf{GPT4o}   & \multicolumn{1}{c|}{LLaVA}   & \multicolumn{1}{c|}{\textbf{GPT4o}}   & \multicolumn{1}{c|}{LLaVA}   & \textbf{GPT4o}   \\ \hline
Checker LLM                          & \multicolumn{1}{c|}{Llama3}   & \multicolumn{1}{c|}{\textbf{Llama3}}   & \multicolumn{1}{c|}{GPT4o}   & \textbf{GPT4o}   & \multicolumn{1}{c|}{Llama3}   & \multicolumn{1}{c|}{\textbf{Llama3}}   & \multicolumn{1}{c|}{GPT4o}   & \textbf{GPT4o}   \\ \hline
Precision (correct over consistent)  & \multicolumn{1}{c|}{73.96\%} & \multicolumn{1}{c|}{\textbf{75.34\%}} & \multicolumn{1}{c|}{76.56\%} & \textbf{77.28\%} & \multicolumn{1}{c|}{73.96\%} & \multicolumn{1}{c|}{\textbf{75.34\%}} & \multicolumn{1}{c|}{76.56\%} & \textbf{77.28\%} \\ \hline
Recall (consistent over correct)     & \multicolumn{1}{c|}{99.34\%} & \multicolumn{1}{c|}{\textbf{100.0\%}} & \multicolumn{1}{c|}{87.22\%} & \textbf{98.63\%} & \multicolumn{1}{c|}{97.9\%}  & \multicolumn{1}{c|}{\textbf{99.84\%}} & \multicolumn{1}{c|}{76.75\%} & \textbf{93.8\%}  \\ \hline
F1 Score                             & \multicolumn{1}{c|}{84.79\%} & \multicolumn{1}{c|}{\textbf{85.93\%}} & \multicolumn{1}{c|}{81.54\%} & \textbf{86.66\%} & \multicolumn{1}{c|}{84.26\%} & \multicolumn{1}{c|}{\textbf{85.87\%}} & \multicolumn{1}{c|}{76.65\%} & \textbf{84.75\%} \\ \hline
Specificity (flagged hallucinations) & \multicolumn{1}{c|}{4.91\%}  & \multicolumn{1}{c|}{\textbf{1.47\%}}  & \multicolumn{1}{c|}{46.44\%} & \textbf{23.18\%} & \multicolumn{1}{c|}{0.93\%}  & \multicolumn{1}{c|}{\textbf{0.99\%}}  & \multicolumn{1}{c|}{26.27\%} & \textbf{11.05\%} \\ \hline
Matthews Correlation Coefficient     & \multicolumn{1}{c|}{0.1418}  & \multicolumn{1}{c|}{\textbf{0.1052}}  & \multicolumn{1}{c|}{0.3713}  & \textbf{0.3728}  & \multicolumn{1}{c|}{-0.0386} & \multicolumn{1}{c|}{\textbf{0.0587}}  & \multicolumn{1}{c|}{0.0303}  & \textbf{0.0790}  \\ \hline

\end{tabular}
}
\label{tab:PREPER_NotOverlookingAgents}
\end{table}

\subsection{Time of day effect on hallucination detection}
\label{sec:TimeofDayEffect}

The Waymo dataset contains three different labels, `Day', `Dawn and dusk', and `Night' to denote the time of the day each image was captured. These label data were extracted to categorize the hallucination detection results to understand the variations in the performance metrics in terms of the time of the day. This section includes tables that present such categorized results retrieved for Waymo images, under the definition of correctness detecting hallucinations: traffic agents in the generated captions that do not appear in the manual annotations. 

Table \ref{tab:DayTime_Hallucinationdetection} showcases the performances of hallucination detection when the captions are generated only for the images captured during daytime. Tab.~\ref{tab:DuskDawnTime_Hallucinationdetection} showcases the performances of hallucination detection when the captions are generated for the Waymo images captured during dawn and dusk. Finally, Tab.~\ref{tab:NightTime_Hallucinationdetection} showcases the performances of hallucination detection when the captions are generated for the Waymo images captured during nighttime.  
\begin{table}[]
\caption{Performance of hallucination detection using Llama3 and GPT-4o for the captions generated by GPT-4o and LLaVA for Waymo images captured during the daytime. Correctness is defined as not hallucinating traffic agents. The performances are compared for the original response $R_1$, before filtering sentences, and for $R'_1$.}
\resizebox{\textwidth}{!}{%
\begin{tabular}{|l|cccc|cccc|}
\hline
\textbf{Dataset: Waymo}              & \multicolumn{4}{l|}{Fixed response $R'_1$}                                                                             & \multicolumn{4}{l|}{Original response $R_1$}                                                                           \\ \hline
Captioner LLM                        & \multicolumn{1}{c|}{LLaVA}   & \multicolumn{1}{c|}{\textbf{GPT4o}}   & \multicolumn{1}{c|}{LLaVA}   & \textbf{GPT4o}   & \multicolumn{1}{c|}{LLaVA}   & \multicolumn{1}{c|}{\textbf{GPT4o}}   & \multicolumn{1}{c|}{LLaVA}   & \textbf{GPT4o}   \\ \hline
Checker LLM                          & \multicolumn{1}{c|}{Llama3}   & \multicolumn{1}{c|}{\textbf{Llama3}}   & \multicolumn{1}{c|}{GPT4o}   & \textbf{GPT4o}   & \multicolumn{1}{c|}{Llama3}   & \multicolumn{1}{c|}{\textbf{Llama3}}   & \multicolumn{1}{c|}{GPT4o}   & \textbf{GPT4o}   \\ \hline
Precision (correct over consistent)  & \multicolumn{1}{c|}{81.74\%} & \multicolumn{1}{c|}{\textbf{93.24\%}} & \multicolumn{1}{c|}{91.26\%} & \textbf{93.95\%} & \multicolumn{1}{c|}{81.45\%} & \multicolumn{1}{c|}{\textbf{93.24\%}} & \multicolumn{1}{c|}{83.29\%} & \textbf{93.37\%} \\ \hline
Recall (consistent over correct)     & \multicolumn{1}{c|}{99.65\%} & \multicolumn{1}{c|}{\textbf{99.71\%}} & \multicolumn{1}{c|}{84.93\%} & \textbf{95.88\%} & \multicolumn{1}{c|}{99.65\%} & \multicolumn{1}{c|}{\textbf{99.71\%}} & \multicolumn{1}{c|}{88.28\%} & \textbf{96.43\%} \\ \hline
F1 Score                             & \multicolumn{1}{c|}{89.81\%} & \multicolumn{1}{c|}{\textbf{96.37\%}} & \multicolumn{1}{c|}{87.98\%} & \textbf{94.91\%} & \multicolumn{1}{c|}{89.63\%} & \multicolumn{1}{c|}{\textbf{96.37\%}} & \multicolumn{1}{c|}{85.71\%} & \textbf{94.88\%} \\ \hline
Specificity (flagged hallucinations) & \multicolumn{1}{c|}{0.0\%}   & \multicolumn{1}{c|}{\textbf{0.0\%}}   & \multicolumn{1}{c|}{8.11\%}  & \textbf{0.0\%}   & \multicolumn{1}{c|}{0.0\%}   & \multicolumn{1}{c|}{\textbf{0.0\%}}   & \multicolumn{1}{c|}{26.14\%} & \textbf{8.0\%}   \\ \hline
Matthews Correlation Coefficient     & \multicolumn{1}{c|}{-0.0254} & \multicolumn{1}{c|}{\textbf{-0.0139}} & \multicolumn{1}{c|}{-0.0540} & \textbf{-0.0499} & \multicolumn{1}{c|}{-0.0256} & \multicolumn{1}{c|}{\textbf{-0.0139}} & \multicolumn{1}{c|}{0.1617}  & \textbf{0.0582}  \\ \hline

\end{tabular}
}
\label{tab:DayTime_Hallucinationdetection}
\end{table}

\begin{table}[]
\caption{Performance of hallucination detection using Llama3 and GPT-4o for the captions generated by GPT-4o and LLaVA for Waymo images captured during dawn and dusk. Correctness is defined as not hallucinating traffic agents. Performances are compared for the original response $R_1$, before filtering sentences, and for $R'_1$.}
\resizebox{\textwidth}{!}{%
\begin{tabular}{|l|cccc|cccc|}
\hline
\textbf{Dataset: Waymo}              & \multicolumn{4}{l|}{Fixed response $R'_1$}                                                                             & \multicolumn{4}{l|}{Original response $R_1$}                                                                           \\ \hline
Captioner LLM                        & \multicolumn{1}{c|}{LLaVA}   & \multicolumn{1}{c|}{\textbf{GPT4o}}   & \multicolumn{1}{c|}{LLaVA}   & \textbf{GPT4o}   & \multicolumn{1}{c|}{LLaVA}   & \multicolumn{1}{c|}{\textbf{GPT4o}}   & \multicolumn{1}{c|}{LLaVA}   & \textbf{GPT4o}   \\ \hline
Checker LLM                          & \multicolumn{1}{c|}{Llama3}   & \multicolumn{1}{c|}{\textbf{Llama3}}   & \multicolumn{1}{c|}{GPT4o}   & \textbf{GPT4o}   & \multicolumn{1}{c|}{Llama3}   & \multicolumn{1}{c|}{\textbf{Llama3}}   & \multicolumn{1}{c|}{GPT4o}   & \textbf{GPT4o}   \\ \hline
Precision (correct over consistent)  & \multicolumn{1}{c|}{86.38\%} & \multicolumn{1}{c|}{\textbf{99.64\%}} & \multicolumn{1}{c|}{95.08\%} & \textbf{100.0\%} & \multicolumn{1}{c|}{86.38\%} & \multicolumn{1}{c|}{\textbf{99.64\%}} & \multicolumn{1}{c|}{90.53\%} & \textbf{99.68\%} \\ \hline
Recall (consistent over correct)     & \multicolumn{1}{c|}{99.11\%} & \multicolumn{1}{c|}{\textbf{100.0\%}} & \multicolumn{1}{c|}{79.68\%} & \textbf{97.21\%} & \multicolumn{1}{c|}{99.11\%} & \multicolumn{1}{c|}{\textbf{100.0\%}} & \multicolumn{1}{c|}{82.13\%} & \textbf{97.2\%}  \\ \hline
F1 Score                             & \multicolumn{1}{c|}{92.31\%} & \multicolumn{1}{c|}{\textbf{99.82\%}} & \multicolumn{1}{c|}{86.7\%}  & \textbf{98.59\%} & \multicolumn{1}{c|}{92.31\%} & \multicolumn{1}{c|}{\textbf{99.82\%}} & \multicolumn{1}{c|}{86.13\%} & \textbf{98.43\%} \\ \hline
Specificity (flagged hallucinations) & \multicolumn{1}{c|}{0.0\%}   & \multicolumn{1}{c|}{\textbf{0.0\%}}   & \multicolumn{1}{c|}{23.53\%} & \textbf{0.0\%}   & \multicolumn{1}{c|}{0.0\%}   & \multicolumn{1}{c|}{\textbf{0.0\%}}   & \multicolumn{1}{c|}{39.02\%} & \textbf{0.0\%}   \\ \hline
Matthews Correlation Coefficient     & \multicolumn{1}{c|}{-0.0348} & \multicolumn{1}{c|}{\textbf{0.0}}     & \multicolumn{1}{c|}{0.0175}  & \textbf{0.0}     & \multicolumn{1}{c|}{-0.0348} & \multicolumn{1}{c|}{\textbf{0.0}}     & \multicolumn{1}{c|}{0.1724}  & \textbf{-0.0094} \\ \hline

\end{tabular}
}
\label{tab:DuskDawnTime_Hallucinationdetection}
\end{table}

\begin{table}[]
\caption{Performance of hallucination detection using Llama3 and GPT-4o for the captions generated by GPT-4o and LLaVA for Waymo images captured during night time. Correctness is defined as not hallucinating traffic agents. The performances are compared for the original response $R_1$, before filtering sentences, and for $R'_1$.}
\resizebox{\textwidth}{!}{%
\begin{tabular}{|l|cccc|cccc|}
\hline
\textbf{Dataset: Waymo}              & \multicolumn{4}{l|}{Fixed response $R'_1$}                                                                             & \multicolumn{4}{l|}{Original response $R_1$}                                                                           \\ \hline
Captioner LLM                        & \multicolumn{1}{c|}{LLaVA}   & \multicolumn{1}{c|}{\textbf{GPT4o}}   & \multicolumn{1}{c|}{LLaVA}   & \textbf{GPT4o}   & \multicolumn{1}{c|}{LLaVA}   & \multicolumn{1}{c|}{\textbf{GPT4o}}   & \multicolumn{1}{c|}{LLaVA}   & \textbf{GPT4o}   \\ \hline
Checker LLM                          & \multicolumn{1}{c|}{Llama3}   & \multicolumn{1}{c|}{\textbf{Llama3}}   & \multicolumn{1}{c|}{GPT4o}   & \textbf{GPT4o}   & \multicolumn{1}{c|}{Llama3}   & \multicolumn{1}{c|}{\textbf{Llama3}}   & \multicolumn{1}{c|}{GPT4o}   & \textbf{GPT4o}   \\ \hline
Precision (correct over consistent)  & \multicolumn{1}{c|}{88.06\%} & \multicolumn{1}{c|}{\textbf{100.0\%}} & \multicolumn{1}{c|}{91.74\%} & \textbf{100.0\%} & \multicolumn{1}{c|}{88.06\%} & \multicolumn{1}{c|}{\textbf{100.0\%}} & \multicolumn{1}{c|}{90.91\%} & \textbf{100.0\%} \\ \hline
Recall (consistent over correct)     & \multicolumn{1}{c|}{98.33\%} & \multicolumn{1}{c|}{\textbf{99.41\%}} & \multicolumn{1}{c|}{74.0\%}  & \textbf{87.59\%} & \multicolumn{1}{c|}{98.33\%} & \multicolumn{1}{c|}{\textbf{99.41\%}} & \multicolumn{1}{c|}{77.46\%} & \textbf{87.59\%} \\ \hline
F1 Score                             & \multicolumn{1}{c|}{92.91\%} & \multicolumn{1}{c|}{\textbf{99.71\%}} & \multicolumn{1}{c|}{81.92\%} & \textbf{93.39\%} & \multicolumn{1}{c|}{92.91\%} & \multicolumn{1}{c|}{\textbf{99.71\%}} & \multicolumn{1}{c|}{83.65\%} & \textbf{93.39\%} \\ \hline
Specificity (flagged hallucinations) & \multicolumn{1}{c|}{0.0\%}   & \multicolumn{1}{c|}{\textbf{0.0\%}}   & \multicolumn{1}{c|}{0.0\%}   & \textbf{0\%}     & \multicolumn{1}{c|}{0.0\%}   & \multicolumn{1}{c|}{\textbf{0.0\%}}   & \multicolumn{1}{c|}{38.89\%} & \textbf{0\%}     \\ \hline
Matthews Correlation Coefficient     & \multicolumn{1}{c|}{-0.0446} & \multicolumn{1}{c|}{\textbf{0.0}}     & \multicolumn{1}{c|}{-0.1465} & \textbf{0.0}     & \multicolumn{1}{c|}{-0.0446} & \multicolumn{1}{c|}{\textbf{0.0}}     & \multicolumn{1}{c|}{0.1203}  & \textbf{0.0}     \\ \hline

\end{tabular}
}
\label{tab:NightTime_Hallucinationdetection}
\end{table}

On the other hand, caption correctness can also be defined as not overlooking any traffic agents, as this is critical for safety in the automotive domain. The performance metrics values for other times of the day following this definition of correctness are included in the supplementary materials and discussed in Section~\ref{sec:AnalysisAndDiscussion}.

\section{Analysis and Discussion}
\label{sec:AnalysisAndDiscussion}

The analysis of the LLM-generated responses revealed that LLaVA and GPT-4o are capable of generating captions consistent with the ground truth labels. Therefore, the application and adaptation of a hallucination detection technique such as SelfCheckGPT was expected to be effective in filtering out errors by the LLM that would be critical in perception-related tasks in the automotive domain.
The performance matrices in Tables~\ref{tab:NOExtra_HallucinationDetection} and~\ref{tab:NotOverlookingAgents} show that the SelfCheckGPT-like filtering process is slightly more effective for GPT4o-generated captions than for LLaVA ones. The performance of this filtering process is however quite varied across the captioner- and checker-LLMs at the sentence level.

In general, the higher recall and precision values recorded for Llama3 under the hallucination detection definition indicate that this checker LLM model is better at correctly identifying non-hallucinated content. GPT-4o reports lower recall values for LLaVA-generated captions, indicating that some non-hallucinated content generated by LLaVA may have been flagged incorrectly as hallucinations. This behavior is not impacted by the sentence-level filtering process, which was introduced to reduce incorrectly flagged sentences from the captions resulting in increasing the trustworthiness of the final caption. Also, the same analysis applies to the performances reported in Tab.~\ref{tab:NotOverlookingAgents} following the definition of correction of not overlooking traffic agents. However, the precision for the ``not overlooking traffic agents'' approach is lower, indicating that the proposed methodology is better at identifying and detecting non-hallucinated content at the expense of missing hallucinations. Hence, the SelfCheckGPT approach and its adaptation can be applied to filter out hallucinations using state-of-the-art LLMs for image captioning tasks for automotive usage scenarios, yet it comes with a price of missing some hallucinations.

The second research question (RQ2) is concerned with the performance differences of SelfCheckGPT and its adaptations based on the two state-of-the-art datasets, given the different traffic scenarios and geographical areas they cover. However, significant deviations were not visible within the recorded results indicating that the main differences in Waymo and PREPER CITY do not pose any impact on the hallucination detection. 

The results generated for the Waymo dataset were analyzed separately to answer the third research question (RQ3). The main motivation was to identify to what extent the performance of SelfCheckGPT and its adaptations are affected by light conditions. Based on the recorded results, the daytime captured images show better results compared to dawn and dusk or nighttime captured images. The higher performance matrices are recorded for daytime captured images for both correctness definitions.


We used the study by Feldt and Magazinius (2010)~\cite{RobertFeldt_ThreatsToValidity} to assess potential threats to the validity of our study. We heuristically designed a specific prompt that aligns with the operational setup of our experiment by restricting the LLMs from generating lengthier sentences. This bears potentially the risk of missing out on an LLM's preferred or more likely way of describing a traffic situation and hence, potentially penalizing an LLM for not spotting a traffic agent even though its synonyms may have spotted them. However, as prompts are still very difficult to systematize, variants may have been more successful. In addition to that, the use of annotations to normalize the distribution of traffic scenarios may not consider the difficulty level, ie., partially occluded traffic agents for example. Here, we may have unknowingly favored one dataset over the other. Furthermore, as highlighted in the experimental design, vehicles dominate the captured traffic scenarios. This bears potentially the risk that more vulnerable road users such as pedestrians and cyclists are insufficiently represented in the experimental sample. Hence, the performance of the adopted SelfCheckGPT approach may vary if a dataset contains many more traffic scenarios with such vulnerable road users. The use of GPT-4o was considered an industrial gold standard. However, at the same time, this LLM is proprietary and hence, may have undergone unnoticed and non-controllable updates during or after our experimentation. This would potentially affect the robustness of our findings. Furthermore, we had no control over the manual annotations of the objects in the Waymo dataset as we directly used the labels provided by the dataset creators, given that some scenarios with inaccurate labels were identified while randomly checking image samples.

\section{Conclusion and Future Work} \label{sec:conclusion}

We have adopted SelfCheckGPT for an automotive application scenario that is relevant for improving perception stacks for ADAS and AD when they may incorporate LLMs or more specific Foundational Models (FMs). We have compared the performance of SelfCheckGPT and its adaptation to spot potential hallucinations and filter them out from the generated description of the vehicle surroundings. This experimental setup was designed and evaluated using the proprietary, cloud-based LLM GPT-4o, and an offline open-source LLM LLaVA. Both LLMs show exemplary performances on image description tasks when prompted thoroughly. We found that GPT-4o was lenient in finding mismatches with many of the captions demonstrating a tendency to flag more captions as hallucinated, which did improve the overall hallucination detection process, but at a large expense of mislabelling non-hallucinated content. The trade-off between precision and recall should be researched further to fine-tune the proposed methodology by reducing the occurrence of mislabelling. Overall, the SelfCheckGPT setup and its adaptation with sentence level filtering improved the overall performance, however the improvement was marginal.

As highlighted in the previous section, thorough attention needs to be given to vulnerable road users such as pedestrians or cyclists. Similarly, it is very important to reduce the amount of overlooked traffic participants in a given usage scenario, which helps in mitigating the risk of potential collisions. Hence, future studies should focus thereon to identifying specifically challenging scenarios for the SelfCheckGPT approach to improve its potential suitability for automotive perception systems.

\section*{Acknowledgments}
This work has been supported by the Swedish Foundation for Strategic Research (SSF), Grant Number FUS21-0004 SAICOM and the Wallenberg AI, Autonomous Systems and Software Program (WASP) funded by the Knut and Alice Wallenberg Foundation. This research has been partially supported by the Swedish Research Council (Diarienummer: 2024-2028).

\bibliographystyle{splncs04}
\bibliography{references}

\end{document}